\documentclass[runningheads]{llncs}
\usepackage{graphicx}

\usepackage{tikz}
\usepackage{comment} 
\usepackage{amsmath,amssymb} 
\usepackage{color}

\usepackage[nosort, nocompress]{cite}
\usepackage{booktabs}
\usepackage[ruled]{algorithm}
\usepackage{algpseudocode}
\usepackage{multirow}
\usepackage{subcaption}

\newcommand{\tnote}[1]{\textcolor{black}{}}

\newcommand{\doublecheck}[1]{\textcolor{black}{#1}}
\newcommand{\cut}[1]{}
\newcommand{\keypoint}[1]{\vspace{0.0cm}\noindent\textbf{#1}~}

\begin{document}
\pagestyle{headings}
\mainmatter
\def\ECCVSubNumber{100}  

\title{Online Meta-Learning for Multi-Source and Semi-Supervised Domain Adaptation} 

\titlerunning{Online Meta-DA}
%
\author{Da Li\inst{1}\orcidID{0000-0002-2101-2989} \and
Timothy Hospedales\inst{1,2}\orcidID{0000-0003-4867-7486} }
\authorrunning{Da Li and Timothy Hospedales}
%
\institute{Samsung AI Center Cambridge, UK \and
The University of Edinburgh, UK.\\
\email{dali.academic@gmail.com; t.hospedales@ed.ac.uk}}
\maketitle

\begin{abstract}
Domain adaptation (DA) is the topical problem of adapting models from labelled source datasets so that they perform well on target datasets where only unlabelled or partially labelled data is available. Many methods have been proposed to address this problem through different ways to minimise the domain shift between source and target datasets. In this paper we take an orthogonal perspective and propose a framework to further enhance performance by meta-learning the initial conditions of existing DA algorithms. This is challenging compared to the more widely considered setting of few-shot meta-learning, due to the length of the computation graph involved. Therefore we propose an online shortest-path meta-learning framework that is both computationally tractable and practically effective for improving DA performance. We present variants for both multi-source unsupervised domain adaptation (MSDA), and semi-supervised domain adaptation (SSDA). Importantly, our approach is agnostic to the base adaptation algorithm, and can be applied to improve many techniques. Experimentally, we demonstrate improvements on classic (DANN) and recent (MCD  and MME) techniques for MSDA and SSDA, and ultimately achieve state of the art results on several DA benchmarks including the largest scale DomainNet.
\end{abstract}

\section{Introduction}

Contemporary deep learning methods now provide excellent performance across a variety of computer vision tasks when ample annotated training data is available. However, this performance often degrades rapidly if models are applied to novel domains with very different data statistics from the training data, which is a problem known as domain shift. Meanwhile, data collection and annotation for every possible domain of application is expensive and sometimes impossible. This challenge has motivated extensive study in the area of domain adaptation (DA), which addresses  training models that work well on a target domain using only unlabelled or partially labelled target data from that domain together with labelled data from a source domain.

\begin{figure}[t]
    \centering
    \includegraphics[width=0.48\linewidth]{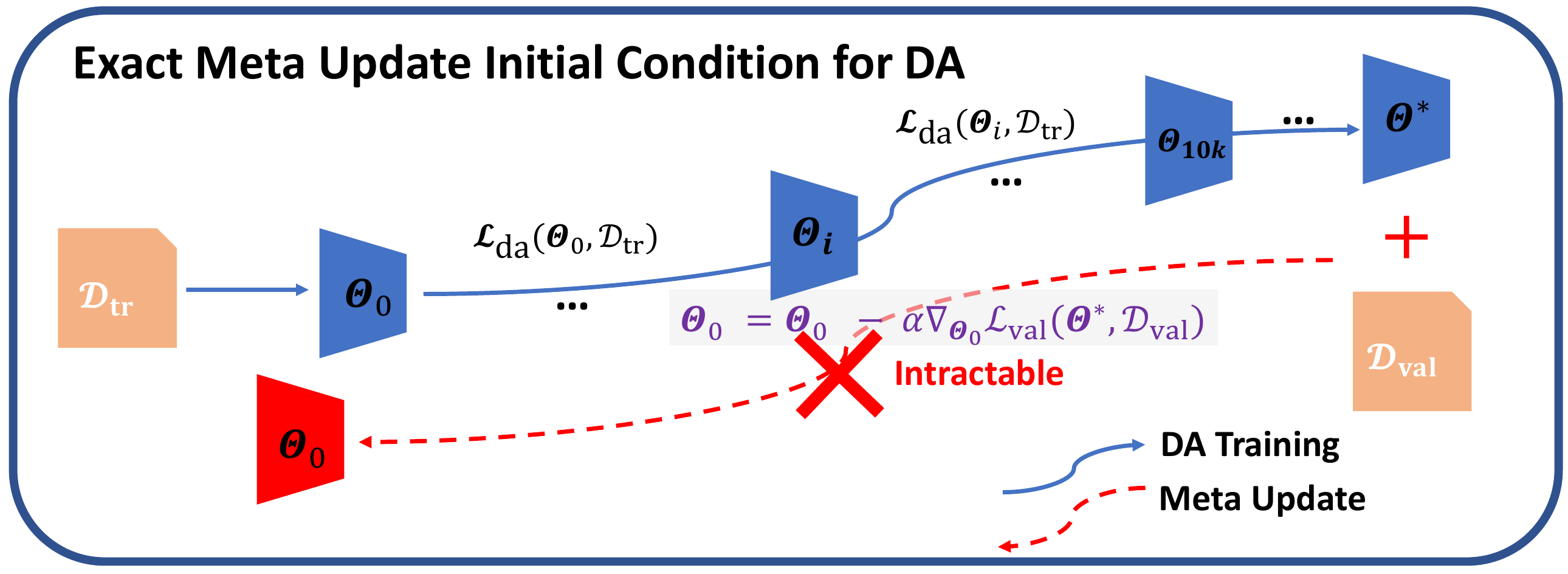}
    \includegraphics[width=0.49\linewidth]{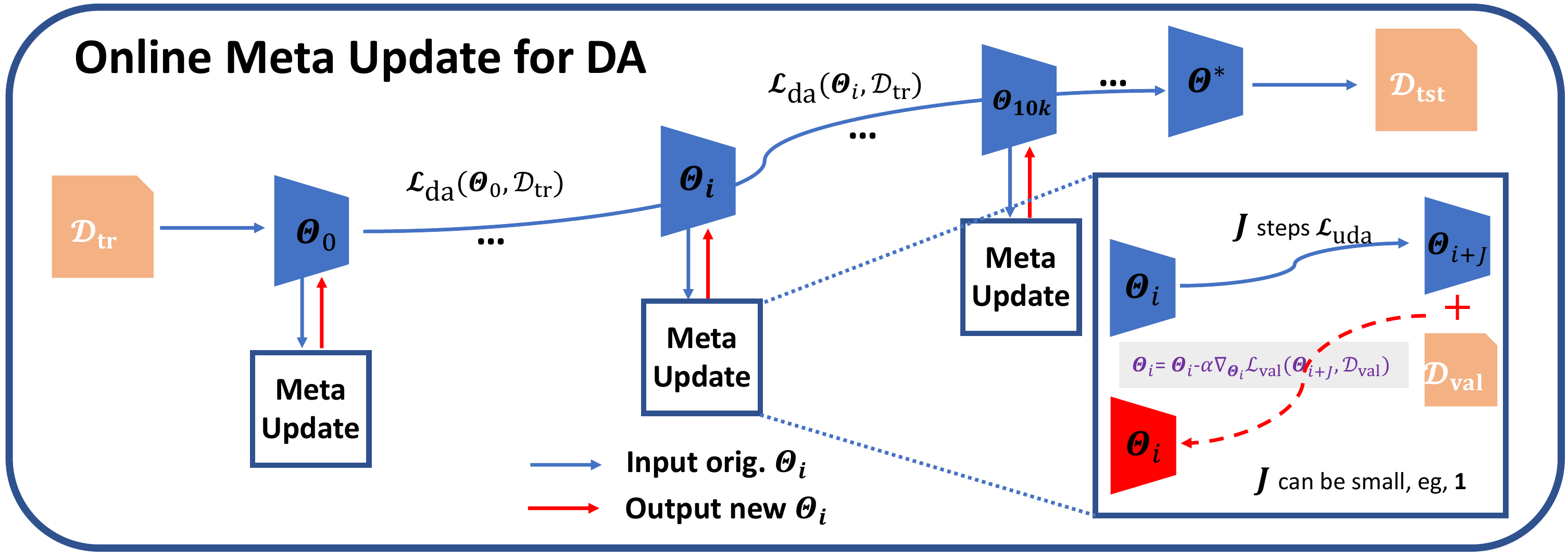}
    \caption{\cut{Illustrative schematics of Online Meta Domain Adaptation. Left: Optimization paths on domain adaptation loss (shading). 
    (Solid line) Vanilla gradient descent on a DA objective from a fixed start point. 
    (Multi-segment line) Online meta-learning iterates meta and gradient descent updates.
    (Two segment line) Sequential meta-learning provides an alternative approximation: update initial condition, then perform gradient descent. }
    \cut{Right: (Top) Exact meta-learning of initial condition with inner loop training DA to convergence is intractable. (Middle) Sequential meta-learning performs meta updates and DA updates sequentially. (Bottom) Online meta-learning alternates between meta-optimization and domain adaptation.}
    Left: Exact meta-learning of initial condition with inner loop training DA to convergence is intractable. 
    Right: Online meta-learning alternates between meta-optimization and domain adaptation.
    }
        \label{fig:illustration}  
\end{figure}

Several variants of the domain adaptation problem have been studied. Single-source domain adaptation (SDA) considers adaptation from a single source domain~\cite{shai2006nipsdomainadaptation, bousmalis2016domain}, while multi-source domain adaptation (MSDA) considers improving cross-domain generalization by aggregating information across multiple sources~\cite{mansour2009domain,tzeng2017adversarial}. Unsupervised domain adaptation (UDA) learns solely from unlabelled data in the target domain \cite{ganin2016domain, saito2018maximum}, while semi-supervised domain adaptation (SSDA) learns from a mixture of labelled and unlabelled target domain data~\cite{donahue2013semi,daume2007easyDA}. The main means of progress has been developing improved methods for aligning representations between source(s) and the target in order to improve generalization. These methods span distribution alignment, for example by maximum mean discrepancy (MMD) \cite{long2015learning, tzeng2014deep}, domain adversarial training \cite{ganin2016domain, saito2018maximum}, and cycle consistent image transformation \cite{NIPS2016_6544, hoffman2018cycada}.

In this paper we adopt a novel research perspective that is complementary to all these existing methods. Rather than proposing a new domain adaptation strategy, we study a meta-learning framework for improving these existing adaptation algorithms. Meta-learning (a.k.a. learning to learn) has a long history~\cite{schmidhuber1992learning,schmidhuber1997shifting}, and has re-surged recently, especially due to its efficacy in improving few-shot deep learning~\cite{vinyals2016matching, ravi2016optimization, finn2017model}. Meta-learning pipelines aim to improve learning by training some aspect of a learning algorithm such a comparison metric~\cite{vinyals2016matching}, model optimizer~\cite{ravi2016optimization} or model initialisation~\cite{finn2017model}, so as to improve outcomes according to some meta-objective such as few-shot learning efficacy \cite{vinyals2016matching, ravi2016optimization, finn2017model} or learning speed \cite{andrychowicz2016learning}. In this paper we provide a first attempt to define a meta-learning framework for improving domain adaptive learning.

We take the perspective of meta-optimizing the \emph{initial condition} \cite{maclaurin2015gradient,finn2017model} of domain adaptive learning\footnote{One may not think of domain adaptation as being sensitive to initial condition, but given the lack of target domain supervision to guide learning, different initialization can lead to a significant 10-15\% difference in accuracy (see Supplementary material).}.  {While there are several facets of algorithms that can be meta-learned such as hyper-parameters \cite{franceschi2018bilevel} and learning rates \cite{li2017metaSGD}; these are \cut{harder to update efficiently, and} somewhat tied to the base learning algorithm (domain adaptive algorithm in our case).} In contrast, our framework is \emph{algorithm agnostic} in that it can be used to improve many existing gradient-based domain adaptation algorithms. 

Furthermore we develop variants for both unsupervised multi-source adaptation, as well as semi-supervised single source adaptation, thus providing broad potential benefit to existing frameworks and application settings. In particular we demonstrate application of our framework to the classic domain adversarial neural network (DANN) \cite{ganin2016domain} algorithm, as well as the recent maxmium-classifier discrepancy (MCD) \cite{saito2018maximum}, and min-max entropy (MME) \cite{saito2019semi} algorithms.

Meta-learning can often be cleanly formalised as a bi-level optimization problem \cite{franceschi2018bilevel,rajeswaran2019metaImplicit}: where an outer loop optimizes the meta-parameter of interest (such as the initial condition in our case) with respect to some meta-loss (such as performance on a validation set); and the inner loop runs the learning algorithm conditioned on the chosen meta-parameter. {This is tricky to apply directly in a domain adaptation scenario however, because: (i) The computation graph of the inner loop is typically long  (unlike the popular few-shot meta-learning setting \cite{finn2017model}), making meta-optimization intractable, and (ii) Especially in unsupervised domain adaptation, there is no labelled data in the target domain to define a supervised learning loss for the outer-loop meta-objective.} We surmount these challenges by proposing a simple, fast and efficient meta-learning strategy based on online 
shortest path gradient descent~\cite{nichol2018first}, and defining meta-learning pipelines suited for meta-optimization of domain adaptation problems. \doublecheck{Although motivated by initial condition learning, our online algorithm ultimately has the interpretation of intermittently performing meta-update(s) of the parameters in order to achieve the best outcome from the following DA updates (Fig~\ref{fig:illustration}).}

Overall, our contributions are: (i) Introducing a meta-learning framework suitable for both multi-source and semi-supervised domain adaptation settings, (ii) We demonstrate the algorithm agnostic nature of our framework through its application to several base domain adaptation methods including MME~\cite{saito2019semi}, DANN~\cite{ganin2016domain} and MCD~\cite{saito2018maximum}, (iii) Applying our meta-learner to these base adaptation methods, we achieve state of the art performance on several MSDA and SSDA benchmarks, including the largest-scale DA benchmark, DomainNet~\cite{peng2019moment}.


\section{Related Work\label{sec:relate-work}}

\keypoint{Single-Source Domain Adaptation}
Single-source unsupervised domain adaptation (SDA) is a well established area \cite{shai2006nipsdomainadaptation,tzeng2014deep,long2015learning,ganin2016domain, long2016unsupervised, bousmalis2016domain,long2017deep, saito2018maximum, Luo_2019_CVPR, Kim_2019_CVPR}. Theoretical results have bound the cross-domain generalization error in terms of  domain divergence \cite{ben2010theory}, and numerous algorithms have been proposed to reduce the divergence between source and target features. Representative approaches include minimising MMD distribution shift \cite{tzeng2014deep, long2015learning} or Wasserstein distance \cite{balaji2019normalized,xu2019wasserstein}, adversarial training~\cite{ganin2016domain, saito2018maximum, lee2019sliced} or alignment by cycle-consistent image translation~\cite{NIPS2016_6544,hoffman2018cycada}. Given the difficulty of SDA, studies have considered exploiting semi-supervised or multi-source  adaptation where possible.

\keypoint{Semi-Supervised Domain Adaptation} This setting assumes that besides the labelled source  and unlabelled target domain data, there are a few labelled samples available in the target domain. Exploiting the few  target labels allows better domain alignment compared to purely unsupervised approaches. Representative approaches are based on regularization \cite{daume2007easyDA}, subspace learning \cite{yao2015semi}, label smoothing \cite{donahue2013semi} and entropy minimisation in the target domain \cite{grandvalet2005semi}. The state of the art method in this area, MME, extends the entropy minimisation idea to adversarial entropy minimisation in a deep network setting \cite{saito2019semi}. 

\keypoint{Multi-Source Domain Adaptation} 
This setting assumes there are multiple labelled source domains for training. In deep learning, simply aggregating all source domains data together often already improves performance due to bigger datasets learning a stronger representation. Theoretical results based on  $\mathcal{H}$-divergence ~\cite{ben2010theory} can still apply after aggregation, and existing SDA methods that attempt to reduce source-target divergence ~\cite{shai2006nipsdomainadaptation,ganin2016domain, saito2018maximum} can be used. Meanwhile, new generalization bounds for  MSDA  have been derived~\cite{zhao2018adversarial, peng2019moment}, which motivate algorithms that align amongst source domains as well as between source and 
target. Nevertheless, practical deep network optimization is non-convex, and the degree of alignment achieved depends on the details of the optimization strategy. Therefore our paradigm of meta-learning the initial condition of optimization is compatible with, and complementary to, all this prior work. 

\keypoint{Meta-Learning for Neural Networks}
Meta-Learning (learning to learn)~\cite{schmidhuber1992learning, thrun1998learntolearn} has experienced a recent resurgence. This has largely been driven by its efficacy for few-shot deep learning via initial condition learning \cite{finn2017model}, optimizer learning \cite{ravi2016optimization} and embedding learning \cite{vinyals2016matching}. More generally it has been applied to improve optimization efficiency~\cite{andrychowicz2016learning},   reinforcement learning~\cite{parisotto2019concurrent}, gradient-based hyperparameter optimization \cite{franceschi2018bilevel} and neural architecture search~\cite{liu2018darts}. We start from the perspective of MAML \cite{finn2017model}, in terms of focusing on learning initial conditions of neural network optimization. However besides the different application (domain adaptation vs few-shot learning), our resulting algorithm is very different as we end up performing meta-optimization online \emph{while} solving the target task rather than in advance of solving it \cite{finn2017model,rajeswaran2019metaImplicit}. {A few recent studies also attempted online meta-learning \cite{xu2018metaGrad,finn2019onlineMeta}, but these are designed specifically to backprop through RL  \cite{xu2018metaGrad} or few-shot supervised  \cite{finn2019onlineMeta} learning. Meta-learning with domain adaptation in the inner loop has not been studied before now. }

In terms of learning with multiple domains a few studies  \cite{Li2018MLDG,balaji2018metareg,dou2019domain} have considered  meta-learning for multi-source \emph{domain generalization}, which evaluates the ability of models to generalise directly without adaptation. In practice these methods use supervised learning in their inner optimization. No meta-learning method has been proposed for the domain adaptation problem addressed here.

\section{Methodology\label{sec:meth}}

\subsection{Background}
\begin{algorithm}[t]
\caption{Meta-Update Initial Condition}\label{alg:bilevel-ml}
\begin{algorithmic}
\State \textbf{Function}~UpdateIC($\Theta_0$, $\mathcal{D}_{\text{tr}}$, $\mathcal{D}_{\text{val}}$, $\mathcal{L}_{\text{inner}}$, $\mathcal{L}_{\text{outer}}$)~:
 \For{$j$ = 1, 2, $\dots$, $J$}\hfill\text{// Inner-level optimization}
 \State $\Theta_j = \Theta_{j-1} - \alpha \nabla_{\Theta_{j-1}}\mathcal{L}_{\text{inner}}(\Theta_{j-1}, (\mathcal{D}_{\text{tr}})_j)$
 \EndFor
 \State $\Theta_0  = \Theta_0 - \alpha\nabla_{\Theta_{0}}\mathcal{L}_{\text{outer}}( \Theta_J, {\mathcal{D}_{\text{val}}})$\hfill\text{//Outer-level step}
 \State \textbf{Output}: $\Theta_0$
\end{algorithmic}
\end{algorithm}


\keypoint{Unsupervised Domain Adaptation} Domain adaptation techniques aim to reduce the domain shift between source domain(s) $\mathcal{D}_S$ and target domain $\mathcal{D}_T$, in order that a model trained on labels from  $\mathcal{D}_S$ performs well when deployed on  $\mathcal{D}_T$. Commonly such algorithms train a model $\Theta$ with a loss  $\mathcal{L}_{\text{uda}}$ that breaks down into a term for supervised learning on the source domain $\mathcal{L}_{\text{sup}}$ and an adaptation loss $\mathcal{L}_{\text{a}}$ that attempts to align the target and source data
\begin{equation}\label{eq:uda}
    \begin{aligned}
          \mathcal{L}_{\text{uda}}(\Theta, \mathcal{D}_{S}, \overline{\mathcal{D}}_{T}) = \mathcal{L}_{\text{sup}}(\Theta, \mathcal{D}_S) + \lambda\mathcal{L}_{\text{a}}(\Theta, \mathcal{D}_{S}, \overline{\mathcal{D}}_{T}).
    \end{aligned}
\end{equation}
\noindent We use notation $\mathcal{D}_S$ and $\overline{\mathcal{D}}_T$ to indicate that the source and target domains contain labelled and unlabelled data respectively. Many existing domain adaptation algorithms~\cite{tzeng2014deep, ganin2016domain, saito2018maximum, peng2019moment}  fit into this template, differing in their definition of the domain alignment loss $\mathcal{L}_{\text{a}}$. In the case of multi-source adaptation~\cite{peng2019moment}, $\mathcal{D}_{S}$ may contain several source domains $\mathcal{D}_{S}=\{ {D}_1, \dots, {D}_N\}$ and the first supervised learning term  $\mathcal{L}_{\text{sup}}$ sums the performance on all of these.

\keypoint{Semi-Supervised Domain Adaptation} In the SSDA setting~\cite{saito2019semi}, we assume a sparse set of labelled target data $\mathcal{D}_T$ is provided along with a large set of unlabelled target data $\overline{\mathcal{D}}_{T}$. The goal is to learn a model that fits both the source and few-shot target labels $\mathcal{L}_\text{sup}$, while also aligning the unlabelled target data to the source with an adaptation loss $\mathcal{L}_a$.
\begin{equation}\label{eq:da-total-ssda}
    \begin{aligned}
 \mathcal{L_{\text{ssda}}}(\Theta,\mathcal{D}_{S},\overline{\mathcal{D}}_{T},\mathcal{D}_T)= & \mathcal{L}_{\text{sup}}(\Theta, \mathcal{D}_{S}) + \mathcal{L}_{\text{sup}}(\Theta, \mathcal{D}_{T}) \\
 & +\lambda \mathcal{L}_{\text{a}}(\Theta, \mathcal{D}_{S}, \overline{\mathcal{D}}_{T}) 
    \end{aligned}
\end{equation}
Several existing algorithms \cite{long2018conditional,saito2019semi,ganin2016domain} fit this template and hence can potentially be optimized by our framework.

\keypoint{Meta Learning Model Initialisation}
The problem of meta-learning the initial condition of an optimization can be seen as a bi-level optimization problem~\cite{franceschi2018bilevel,rajeswaran2019metaImplicit}. In this view there is a standard task-specific (inner) algorithm of interest whose initial condition we wish to optimize, and an outer-level meta-algorithm that optimizes that initial condition. This setup can be described as
\begin{equation}\label{eq:meta-learning}
\begin{aligned}
    \Theta & =\underbrace{\underset{\Theta}{\operatorname{argmin} } ~~ \mathcal{L}_{\text{outer}}( \overbrace{\mathcal{L}_{\text{inner}}(\Theta, \mathcal{D}_{\text{tr}})}^{\text{Inner-level}}, \mathcal{D}_{\text{val}})}_{\text{Outer-level}} 
\end{aligned}
\end{equation}
\noindent where $\mathcal{L}_{\text{inner}}(\Theta,\mathcal{D}_{\text{tr}})$ denotes the standard loss of the base task-specific algorithm on its training set.  $\mathcal{L}_{\text{outer}}(\Theta^*,\mathcal{D}_\text{val})$ denotes the validation set loss \emph{after} $\mathcal{L}_{\text{inner}}$ has been optimized, ($\Theta^*=\operatorname{argmin}\mathcal{L}_\text{inner}$), when starting from the initial condition set by the outer optimization. The overall goal in  Eq.~\ref{eq:meta-learning} above is thus to set the initial condition of base algorithm $\mathcal{L}_{\text{inner}}$ such that it achieves minimum loss on the validation set. When both losses are  differentiable we can in principle solve Eq.~\ref{eq:meta-learning}  by taking gradient steps on $\mathcal{L}_\text{outer}$ as shown in Algorithm~\ref{alg:bilevel-ml}. However, such exact meta-learning requires backpropagating through the path of the inner optimization, which is costly and inaccurate for a long computation graph.



 \begin{algorithm}[t]
\caption{Meta-Update Initial Condition: SPG}\label{alg:bilevel-ml-short}
\begin{algorithmic}
\State \textbf{Function}~~UpdateIC($\Theta_0$, $\mathcal{D}_{\text{tr}}$, $\mathcal{D}_{\text{val}}$, $\mathcal{L}_{\text{inner}}$, $\mathcal{L}_{\text{outer}}$)~:
 \State $\tilde{\Theta}_{0} = \operatorname{copy}(\Theta_{0})$
 \For{$j$ = 1, 2, $\dots$, $J$}\hfill\text{// Inner-level optimization}
 \State $\tilde{\Theta}_{j} = \tilde{\Theta}_{j-1} - \alpha \nabla_{\tilde{\Theta}_{j-1}}\mathcal{L}_{\text{inner}}(\tilde{\Theta}_{j-1}, (\mathcal{D}_{\text{tr}})_j)$
 \EndFor
 \State $\nabla^{\text{short}}_{\Theta_{0}} = \Theta_{0} - \tilde{\Theta}_{J}$\hfill\text{//Outer-level step}
 \State $\Theta_0  = \Theta_0 - \alpha\nabla_{\Theta_{0}}\mathcal{L}_{\text{outer}}( \Theta_0-\nabla^{\text{short}}_{\Theta_{0}}, {\mathcal{D}_{\text{val}}})$
 \State \textbf{Output}: $\Theta_0$
\end{algorithmic}
\end{algorithm}

\subsection{Meta-Learning for Domain Adaptation}

\keypoint{Overview} 
For meta domain adaptation, we would like to instantiate the initial condition learning idea summarised earlier in Eq.~\ref{eq:meta-learning} in order to initialize popular domain adaptation algorithms such as \cite{ganin2016domain,saito2018maximum,saito2019semi} that can be represented as problems in the form of Eqs.~\ref{eq:uda} and \ref{eq:da-total-ssda}, so as to to maximise the resulting performance in the target domain upon deployment. To this end we will introduce in the following section appropriate definitions of the inner and outer tasks, as well as a tractable optimization strategy. 


\keypoint{Multi-Source Domain Adaptation}
Suppose we have an adequate algorithm to optimize for initial conditions as required in Eq.~\ref{eq:meta-learning}. How could we apply this idea to multi-source unsupervised domain adaptation setting, given that there is no target domain training data to take the role of $\mathcal{D}_{\text{val}}$ in providing the metric for outer loop optimization of the initial condition? Our idea is that in the \emph{multi-source} domain adaptation setting, we can split available source domains into disjoint meta-training and meta-testing domains $\mathcal{D}_S=\mathcal{D}_S^{\text{mtr}}\cup\mathcal{D}_S^{\text{mte}}$, where we actually have labels for both. Now we can let $\mathcal{L}_\text{inner}$ be an unsupervised domain  method  \cite{ganin2016domain,saito2018maximum} $\mathcal{L}_\text{inner}:=\mathcal{L}_\text{uda}$, and ask it to adapt from meta-train to the unlabelled meta-test domain. In the outer loop, we can then use the labels of the meta-test domain as a validation set to evaluate the adaptation performance via a supervised loss $\mathcal{L}_\text{outer}:=\mathcal{L}_\text{sup}$, such as cross-entropy. 
Thus we aim to find an initial condition for our base domain adaptation method $\mathcal{L}_\text{uda}$ that enables it to adapt effectively between source domains
\begin{equation}
    \Theta_0 = \underset{\Theta_0}{\operatorname{argmin}}\hspace{-0.4cm} \sum_{\mathcal{D}_S^{\text{mtr}},\mathcal{D}_S^{\text{mte}}\sim \mathcal{D}_S}\mathcal{L}_\text{sup}(\mathcal{L}_\text{uda}(\mathcal{D}_S^{\text{mtr}},\overline{\mathcal{D}}_S^{\text{mte}};\Theta_0),\mathcal{D}_S^{\text{mte}})\label{eq:simpleMetamsda}
\end{equation}
\noindent where we use $\mathcal{L}(\cdot;\Theta_0)$ to denote optimizing a loss from starting point $\Theta_0$. This initial condition could be optimized by taking gradient descent steps on the outer supervised loss using $\operatorname{UpdateIC}$ from Algorithm~\ref{alg:bilevel-ml}. The resulting $\Theta_0$ is suited to adapting between all source domains hence should also be good for adapting to the target domain. Thus we would finally instantiate the same UDA algorithm using the learned initial condition, but this time between the full set of source domains, and the true unlabelled target domain $\overline{\mathcal{D}}_T$. 
\begin{equation}
    \Theta = \underset{\Theta}{\operatorname{argmin}}~~ \mathcal{L}_\text{uda}(\mathcal{D}_{S},\overline{\mathcal{D}}_{T};\Theta_0)\label{eq:simplemsda}
\end{equation} 
\tnote{What's the best way to indicate setting inital condition in this kind of notation? }\tnote{Where is the best place to introduce the notation of MLMI() as an algorithm to solve, or take one step on the Eq 4 optimization? Do we need to distinguish between MLMI as a solver, and MLMI as a step update? If so what's the best way to show which one we mean when?}

\keypoint{An Online Solution}
While conceptually simple, the problem with the direct approach above is that it requires completing domain-adaptive training multiple times in the inner optimization. Instead we propose to perform \emph{online} meta-learning  \cite{xu2018metaGrad,zheng2018intrinsicRewardPG} by alternating between steps of meta-optimization of Eq.~\ref{eq:simpleMetamsda} and steps on the final unsupervised domain adaptation problem in Eq.~\ref{eq:simplemsda}. That is, we iterate
\begin{equation}
    \begin{aligned}\label{eq:mlmi-msda}
    \displaystyle
        \Theta &= \operatorname{UpdateIC} (\Theta, (\mathcal{D}_S^{\text{mtr}})\cup(\overline{\mathcal{D}}_S^{\text{mte}}), (\mathcal{D}_S^{\text{mte}}), \mathcal{L}_{\text{uda}}, \mathcal{L}_{\text{sup}} )\\
        \Theta &= \Theta_{} - \alpha \nabla_{\Theta_{}}\mathcal{L}_{\text{uda}}(\Theta_{}, (\mathcal{D}_S),(\overline{\mathcal{D}}_T) )
    \end{aligned}
\end{equation}
\noindent where $(\mathcal{D})$ denotes minibatch sampling from the corresponding dataset, and we call $\operatorname{UpdateIC}(\cdot)$ with a small number of inner-loop optimizations such as $J=1$. 

Our method,  summarised in Figure~\ref{fig:illustration} and  Algorithm~\ref{alg:meta-msda-short}, performs meta-learning online, by simultaneously solving the meta-objective and the target task. \cut{Online meta-learning could be unintuitive here as our meta-parameter of interest is the initial condition.}  
\doublecheck{It translates to tuning the initial condition between taking optimization steps on the target DA task. This avoids the intractability and instability of backpropagating through the long computational graph in the exact approach that meta-optimizes $\Theta_0$ to completion before doing DA. Online meta-learning is also potentially advantageous in practice due to improving optimization throughout training rather than only at the start -- c.f. the vanilla exact method, where the impact of the initial condition on the final outcome is very indirect.}


\keypoint{Semi-Supervised Domain Adaptation} We next consider how to adapt the ideas introduced previously to the semi-supervised domain adaptation setting. In the MSDA setting above, we divided source domains into meta-train and meta-test, used unlabeled data from meta test to drive adaptation, and then used meta-test labels to validate the adaptation performance. In SSDA we do not have multiple source domains with which to use such a meta-train/meta-test split strategy, but we do have a small amount of labeled data in the target domain that we can use to validate adaptation performance and drive initial condition optimization. By analogy to Eq.~\ref{eq:simpleMetamsda}, we can aim to find the initial condition for the unsupervised component $\mathcal{L}_\text{uda}$ of an SSDA method. But now we can use the few labelled examples $\mathcal{D}_T$ to validate the adaptation in the outer loop. 
\begin{equation}
    \Theta_0 = \underset{\Theta_0}{\operatorname{argmin}} \sum \mathcal{L}_\text{sup}(\mathcal{L}_\text{uda}(\mathcal{D}_{\text{S}},\overline{\mathcal{D}}_{T}; \Theta_0),\mathcal{D}_{T})\label{eq:simpleMetassda}
\end{equation}
\noindent The learned initial condition can then be used to instantiate the final semi-supervised domain adaptive training. 
\begin{equation}
    \Theta =\underset{\Theta}{\operatorname{argmin}}~~  \mathcal{L_{\text{ssda}}}(\Theta,\mathcal{D}_{S},\overline{\mathcal{D}}_{T},\mathcal{D}_T;\Theta_0)\label{eq:simplessda}
\end{equation}
\keypoint{An Online Solution} The exact meta SSDA approach above suffers from the same limitations as exact MetaMSDA. So we again apply online meta-learning by iterating between meta-optimization of Eq.~\ref{eq:simpleMetassda} and the final supervised domain adaptation problem of Eq.~\ref{eq:simplessda}.
\begin{equation}
    \begin{aligned}
    \displaystyle
        \Theta &= \operatorname{UpdateIC} (\Theta, (\mathcal{D}_S)\cup(\overline{\mathcal{D}}_T), (\mathcal{D}_T), \mathcal{L}_{\text{uda}}, \mathcal{L}_{\text{sup}} )\\
        \Theta &= \Theta -\alpha \nabla_\Theta \mathcal{L_{\text{ssda}}}(\Theta,\mathcal{D}_{S},\overline{\mathcal{D}}_{T},\mathcal{D}_T;\Theta)\label{eq:simplessda}
    \end{aligned}
\end{equation}
The final procedure is summarized in Algorithm~\ref{alg:meta-ssda-short}.

\subsection{Shortest Path optimization}
\begin{algorithm}[t]
\caption{Online Meta Learning: Multi-Source DA}\label{alg:meta-msda-short}
\begin{algorithmic}
\State \textbf{Input}: N source domains $\mathcal{D}_S = [D_1, D_2, \dots, D_N]$  and  unlabelled target domain $\overline{\mathcal{D}}_T$.
 \State \textbf{Initialise}: Model parameters $\Theta$, learning rate $\alpha$, task loss $\mathcal{L}_{\text{sup}}$, UDA method $\mathcal{L}_{\text{uda}}$.
 \For{$i$ = 1, 2, $\dots$, $I$}
 \State $\displaystyle{\Theta_{} = \textbf{UpdateIC}(\Theta_{}, \mathcal{D}_S^{\text{mtr}}\cup\overline{\mathcal{D}}_S^{\text{mte}}, \mathcal{D}_S^{\text{mte}}, \mathcal{L}_{\text{uda}}, \mathcal{L}_{\text{sup}}}$)
 \For{$s$ = 1, 2, $\dots$, $S$}
 \State $\Theta = \Theta_{} - \alpha \nabla_{\Theta_{}}\mathcal{L}_{\text{uda}}(\Theta_{}, (\mathcal{D}_S)_i,(\overline{\mathcal{D}}_T)_i )$\hfill\text{// Domain Adaptation Training}
 \EndFor
 \EndFor
 \State \textbf{Output}: $\Theta$
\end{algorithmic}
\end{algorithm}
\keypoint{Meta-Learning Model Initialisation} 
As described so far, our meta-learning approach to domain adaptation relies on the ability meta-optimize initial conditions using gradient descent steps as described in Algorithm~\ref{alg:bilevel-ml}. Such steps evaluate a meta-gradient that depends on the parameter $\Theta^*$ output by the base domain adaptation algorithm
\begin{equation}\label{eq:outer-orig0}
\begin{aligned}
      \Theta_0 = \Theta_0 - \alpha\overbrace{\nabla_{\Theta}\mathcal{L}_{\text{sup}}(\Theta_{}^* , \mathcal{D}_{\text{val}})}^{\text{Meta Gradient}}
\end{aligned}
\end{equation}
Evaluating the meta-gradient directly is impractical because: (i) The inner loop that runs the base domain adaptation algorithm may take multiple gradient descent iterations $j=1\dots J$. This will trigger a large chain of higher-order gradients  $\nabla_{\Theta_{0}}\mathcal{L}_{\text{inner}}(\cdot), \dots,\nabla_{\Theta_{J-1}}\mathcal{L}_{\text{inner}}(\cdot)$\cut{ when the outer loop performs one update}. (ii) More fundamentally, several state of the art domain adaptation algorithms \cite{saito2018maximum,saito2019semi} use  multiple optimization steps when making updates on $\mathcal{L}_\text{inner}$. For example, to adversarially train the deep feature extractor and classifier modules of the model in $\Theta$. Taking gradient steps on $\mathcal{L}_\text{outer}(\Theta^*)$ thus triggers higher-order gradients, even if one only takes a single step $J=1$ of domain adaptation optimization. 

\keypoint{Shortest Path optimization} 
To obtain the meta gradient in Eq.~\ref{eq:outer-orig0} efficiently, we use shortest-path gradient (SPG)~\cite{nichol2018first}. 
Before optimising the innner loop, we copy parameters $\Theta_0$ as $\tilde{\Theta}_{0}$ and use $\tilde{\Theta}_{0}$ in the inner-level algorithm.  Then, after finishing the inner loop we get the shortest-path gradient between $\tilde{\Theta}_{J}$ and $\Theta_0$.
\begin{equation}\label{eq:short-gradient}
    \begin{aligned}
    \nabla_{\Theta_0}^{\text{short}} = \Theta_0 -  \tilde{\Theta}_{J}
    \end{aligned}
\end{equation}
\noindent Each meta-gradient step (Eq.~\ref{eq:outer-orig0}) is then approximated as
\begin{equation}\label{eq:outer-short}
\begin{aligned}
      \Theta_0 &= \Theta_0 - \alpha \nabla_{\Theta_0}\mathcal{L}_{\text{sup}}(\Theta_0 - \nabla_{\Theta_0}^{\text{short}} , \mathcal{D}_{\text{val}}) \\
\end{aligned}
\end{equation}
\keypoint{Summary} We now have an efficient implementation of $\operatorname{UpdateIC}$ for updating initial conditions as summarised in Algorithm~\ref{alg:bilevel-ml-short}. This shortest path approximation has the advantage of allowing efficient initial condition updates both for multiple iterations of inner loop optimization $J>1$, as well as for inner loop domain adaptation algorithms that use multiple steps \cite{saito2018maximum,saito2019semi}. We use this implementation for the MSDA and SSDA methods in Algorithms~\ref{alg:meta-msda-short} and~\ref{alg:meta-ssda-short}.

\begin{algorithm}[t]
\caption{Online Meta Learning: Semi-Supervised DA}\label{alg:meta-ssda-short}
\begin{algorithmic}
\State \textbf{Input}: N source domains $\mathcal{D}_S = [D_1, D_2, \dots, D_N]$, labelled and unlabelled target domain data $\mathcal{D}_T$ and $\overline{\mathcal{D}}_T$.
 \State \textbf{Initialise}: Model params $\Theta$, learning rate $\alpha$, task loss $\mathcal{L}_{\text{sup}}$, UDA method $\mathcal{L}_{\text{uda}}$.
 \For{$i$ = 1, 2, $\dots$, $I$}
 \State $\displaystyle{\Theta_{} = \textbf{UpdateIC}(\Theta_{}, \mathcal{D}_S\cup\overline{\mathcal{D}}_T, \mathcal{D}_T, \mathcal{L}_{\text{uda}}, \mathcal{L}_{\text{sup} })}$
 \For{$s$ = 1, 2, $\dots$, $S$}
 \State $\scriptstyle{ \Theta = \Theta_{} - \alpha \nabla_{\Theta_{}}(\mathcal{L}_{\text{sup}}(\Theta, (\mathcal{D}_T)_i) +\mathcal{L}_{\text{uda}}(\Theta_{}, (\mathcal{D}_S)_i,(\overline{\mathcal{D}}_T)_i )})$\hfill\text{// Domain Adaptation Training}
 \EndFor
 \EndFor
 \State \textbf{Output}: $\Theta$
\end{algorithmic}
\end{algorithm}


\section{Experiments\label{sec:exp}}
\keypoint{Datasets}
We evaluate our method on several multi-source domain adaptation benchmarks including PACS \cite{li2017pacsDG}, 
Office-Home~\cite{venkateswara2017Deep} and DomainNet \cite{peng2019moment}; as well as on the semi-supervised setting of Office-Home and DomainNet. \\
\keypoint{Base Domain Adaptation Algorithms and Ablation} Our Meta-DA framework is designed to complement existing base domain adaptation algorithms. We evaluate it in conjunction with Domain Adversarial Neural Networks (DANN, \cite{ganin2016domain}) -- as a representative classic approach to deep domain adaptation; as well as Maximum Classifier Discrepancy (MCD, \cite{saito2018maximum}) and MinMax Entropy (MME, \cite{saito2019semi}) -- as examples of state of the art multi-source and semi-supervised domain adaptation algorithms respectively. Our goal is to evaluate whether our Meta-DA framework can improve these base learners. We note that the MCD algorithm has two variants: (1) A multi-step variant that alternates between updating the classifiers and several steps of updating the feature extractor and (2) A one-step variant that uses a gradient reversal \cite{ganin2016domain} layer so that classifier and feature extractor can be updated in a single gradient step. We evaluate both of these. \doublecheck{Sequential Meta-Learning: As an ablation, we also consider an alternative fast meta-learning approach that performs all meta-updates at the start of learning, before doing DA; rather than performing meta-updates online with DA as in our proposed Meta-DA algorithms.}

\begin{table}[t]
    \centering
    \resizebox{0.70\linewidth}{!}{
    \begin{tabular}{l|cccc|c}
    \toprule
    Method & C,P,S $\mapsto$ A & A,P,S $\mapsto$ C & A,C,S $\mapsto$ P & A,C,P $\mapsto$ S & Ave.  \\ \midrule
        Source-only & 77.85 & 74.86 & 95.73 & 67.74  & 79.05 \\
        DIAL~\cite{mancini2018boosting} & 87.30 & 85.50 & 97.00 & 66.80 & 84.15 \\
        DDiscovery~\cite{mancini2018boosting} & \textbf{87.70} & \textbf{86.90} & 97.00 & 69.60 & 85.30 \\
        JiGen~\cite{carlucci2019domain} & 84.88 & 81.07 & \textbf{97.96} & \textbf{79.05} & 85.74 \\
        \midrule
        DANN~\cite{ganin2016domain} & 84.77&83.83	&96.29	&69.61 &83.62\\
        Meta-DANN (Ours) &87.30 &84.90 &96.89&73.22 &85.58\\
        \midrule
        MCD (n=4)~\cite{saito2018maximum} & 86.32 & 84.51 & \textcolor{red}{97.31}& 71.01 & 84.79 \\
                Meta-MCD (n=4) (Ours) & \textcolor{red}{87.40} & \textcolor{red}{86.18} & 97.13 & \textcolor{red}{78.26} & \textbf{87.24} \\
                \midrule
        MCD (os)~\cite{saito2018maximum} & 85.99 & 82.89 & 97.24 & 74.49 & 85.15 \\
        Meta-MCD (os) (Ours) & 86.67 & 84.94 & 96.23 & 77.70 & \textcolor{red}{86.39} \\
        \bottomrule
    \end{tabular}
    }
    \caption{Multi-Source DA results on PACS. Bold: Best. Red: Second Best.}
    \label{tab:pacs-da}
     \vspace{-10pt}
\end{table}
\begin{table}[t]
    \centering
    \resizebox{0.7\linewidth}{!}{
    \begin{tabular}{l|cccc|c}
    \toprule
    Method & C,P,R $\mapsto$ A & A,P,R $\mapsto$ C & A,C,R $\mapsto$ P & A,C,P $\mapsto$ R & Ave. \\
    \midrule
        Source-only & 67.04	&56.04&	80.74&82.86	&71.67  \\
        DSBN~\cite{chang2019domain} & -& -& -& 83.00&- \\
        M$^3$SDA-$\beta$~\cite{peng2019moment} &67.20&58.58&79.05&81.18&71.50\\
        \midrule
        DANN~\cite{ganin2016domain} &68.23 &58.90 & 79.70 & 83.08 & 72.48\\
        Meta-DANN (Ours) & \textbf{70.62}&	59.13&	80.24&	82.79&	73.20\\
        \midrule
        MCD ~\cite{saito2018maximum}  & 69.84 &59.84&	80.92&82.67&73.32  \\
        Meta-MCD (Ours) & 70.21&\textbf{	60.50}&	\textbf{81.17}&	\textbf{83.43}&	\textbf{73.83} \\
        \bottomrule
    \end{tabular}
    }
    \caption{Multi-Source Domain adaptation on Office-Home.}
    \label{tab:office-home}
\end{table}

\subsection{Multi-Source Domain Adaptation}
\subsubsection{PACS:}
\keypoint{Dataset} PACS \cite{li2017pacsDG} was initially proposed for domain generalization and had been subsequently been re-purposed \cite{mancini2018boosting,carlucci2019domain} for  multi-source domain adaptation. This dataset has four diverse domains: (A)rt painting, (C)artoon, (P)hoto and (S)ketch with seven object categories `dog', `elephant', `giraffe', `guitar', `house', `horse' and `person' with 9991 images in total. 
\keypoint{Setting} We follow the setting in~\cite{carlucci2019domain} and perform leave-one-domain out evaluation, setting each domain as the adaptation target in turn. As per~\cite{carlucci2019domain}, we use the ImageNet pre-trained ResNet-18 as our feature extractor for fair comparison. We train with M-SGD (batch size=32, learning rate=$2\times10^{-3}$, momentum=0.9, weight decay=$10^{-4}$). All the models are trained for 5k iterations before testing.
\keypoint{Results} From the results in Table~\ref{tab:pacs-da}, we can see that: (i) Several recent methods with published results on PACS achieve similar performance, with JiGen \cite{carlucci2019domain} performing best. We additionally evaluate DANN and MCD including one-step MCD (os) and multi-step MCD (n=4) variants, and find that one-step MCD performs similarly to JiGen. (ii) Applying our Meta-DA framework to DANN and MCD boosts all three base domain adaptation methods by $1.96\%$, $2.5\%$ and $1.2\%$ respectively. (iii) In particular, our Meta-MCD surpasses the previous state of the art performance set by JiGen. Together these results show the broad applicability and absolute efficacy of our method. {Based on these results we focus on the better performing single-step MCD in the following evaluations.}

\begin{table*}[t]
    \centering
    \resizebox{1.0\linewidth}{!}{
    \begin{tabular}{cl|cccccc|c}
    \toprule
   & \multirow{2}{*}{ Method} & inf,pnt,qdr, & clp,pnt,qdr, & clp,inf,qdr, & clp,inf,qdr & clp,inf,qdr, & clp,inf,qdr, & \multirow{2}{*}{Ave.} \\
   & & rel,skt $\mapsto$ clp & rel,skt $\mapsto$ inf &rel,skt $\mapsto$ pnt & rel,skt  $\mapsto$ qdr & qdr,skt  $\mapsto$ rel & qdr,rel  $\mapsto$ skt  \\
   \midrule
  \parbox[t]{2mm}{\multirow{10}{*}{\rotatebox[origin=c]{90}{Various Backbones \cite{peng2019moment}}}}& Source-only & 47.6$\pm$0.52	&13.0$\pm$0.41	&38.1$\pm$0.45&	13.3$\pm$0.39	&51.9$\pm$0.85&	33.7$\pm$0.54	&32.9$\pm$0.54\\
   &DAN~\cite{long2015learning} &	45.4$\pm$0.49	&12.8$\pm$0.86	&36.2$\pm$0.58	&15.3$\pm$0.37	&48.6$\pm$0.72&	34.0$\pm$0.54	&32.1$\pm$0.59\\
&RTN~\cite{long2016unsupervised}&	44.2$\pm$0.57	&12.6$\pm$0.73&	35.3$\pm$0.59&	14.6$\pm$0.76&	48.4$\pm$0.67&	31.7$\pm$0.73&	31.1$\pm$0.68\\
&JAN~\cite{long2017deep}	&	40.9$\pm$0.43&	11.1$\pm$0.61	&35.4$\pm$0.50&	12.1$\pm$0.67&	45.8$\pm$0.59&	32.3$\pm$0.63&	29.6$\pm$0.57\\
&DANN~\cite{ganin2016domain}	&	45.5$\pm$0.59&	13.1$\pm$0.72&	37.0$\pm$0.69&	13.2$\pm$0.77&	48.9$\pm$0.65&	31.8$\pm$0.62&	32.6$\pm$0.68\\
&ADDA~\cite{tzeng2017adversarial}&	47.5$\pm$0.76&	11.4$\pm$0.67&	36.7$\pm$0.53&	14.7$\pm$0.50	&49.1$\pm$0.82&	33.5$\pm$0.49&	32.2$\pm$0.63\\
&SE~\cite{french2017self}&24.7$\pm$0.32	& ~~3.9$\pm$0.47&	12.7$\pm$0.35&	~~7.1$\pm$0.46&	22.8$\pm$0.51	& ~~9.1$\pm$0.49&	16.1$\pm$0.43\\
&MCD~\cite{saito2018maximum}	&	54.3$\pm$0.64	&22.1$\pm$0.70&	45.7$\pm$0.63&	~~7.6$\pm$0.49&	58.4$\pm$0.65&	43.5$\pm$0.57&	38.5$\pm$0.61\\
&DCTN~\cite{xu2018deep}	&48.6$\pm$0.73	&23.5$\pm$0.59&	48.8$\pm$0.63&	~~7.2$\pm$0.46&	53.5$\pm$0.56&	47.3$\pm$0.47&	38.2$\pm$0.57\\
&M$^3$SDA-$\beta$~\cite{peng2019moment}	&	58.6$\pm$0.53	&26.0$\pm$0.89&	52.3$\pm$0.55&	~~6.3$\pm$0.58&	62.7$\pm$0.51&	49.5$\pm$0.76&	42.6$\pm$0.64\\
    \midrule
    \parbox[t]{2mm}{\multirow{5}{*}{\rotatebox[origin=c]{90}{ResNet-18}}}&   Source-only & 56.58$\pm$0.16&	18.97$\pm$0.10&	45.95$\pm$0.16&	11.52$\pm$0.15&	60.79$\pm$0.17&	43.70$\pm$0.03&39.58$\pm$0.09   \\
       & DANN~\cite{ganin2016domain} & 56.34$\pm$0.12&	18.66$\pm$0.09&	47.09$\pm$0.08&	12.27$\pm$0.12&	61.34$\pm$0.07&	45.26$\pm$0.34&	40.16$\pm$0.12 \\
        & Meta-DANN (Ours) & 57.26$\pm$0.17&	\textbf{19.24$\pm$0.09}&	47.29$\pm$0.16&	13.38$\pm$0.15&	61.21$\pm$0.13&	45.53$\pm$0.17&	40.65$\pm$0.04\\
       & MCD~\cite{saito2018maximum}  & 57.64$\pm$0.28 &	18.71$\pm$0.10&	\textbf{47.82$\pm$0.09}&	12.64$\pm$0.16&	\textbf{61.69$\pm$0.10}&	45.61$\pm$0.01&	40.69$\pm$0.05 \\
       & Meta-MCD (Ours) & \textbf{58.37$\pm$0.21}&	19.09$\pm$0.08&	47.63$\pm$0.12&	\textbf{13.70$\pm$0.14}&	61.30$\pm$0.18&	\textbf{45.90$\pm$0.18}&	\textbf{41.00$\pm$0.05} \\
      
       \midrule
     \parbox[t]{2mm}{\multirow{5}{*}{\rotatebox[origin=c]{90}{ResNet-34}}}&   Source-only & 61.50$\pm$0.06 &	21.10$\pm$0.07&	49.13$\pm$0.06&	13.03$\pm$0.18&	64.14$\pm$0.10&	48.19$\pm$0.12&	42.85$\pm$0.05 \\
       & DANN~\cite{ganin2016domain} & 60.95$\pm$0.05&	20.91$\pm$0.11&	50.35$\pm$0.08&	14.53$\pm$0.06&	64.73$\pm$0.02&	49.88$\pm$0.27&	43.56$\pm$0.04 \\
       & Meta-DANN (Ours)& 61.39$\pm$0.03&	\textbf{21.53$\pm$0.14}&	50.49$\pm$0.29&	15.31$\pm$0.28&	64.33$\pm$0.09&	49.87$\pm$0.25&	43.82$\pm$0.07 \\
       & MCD~\cite{saito2018maximum}  &  62.21$\pm$0.12&	20.49$\pm$0.08&	\textbf{50.87$\pm$0.10}&	14.66$\pm$0.30&	\textbf{64.78$\pm$0.06}&	50.10$\pm$0.11&	43.85$\pm$0.05\\
       & Meta-MCD (Ours) & \textbf{62.81$\pm$0.22}&	21.37$\pm$0.07&	50.53$\pm$0.08&	\textbf{15.47$\pm$0.22}&	64.58$\pm$0.16&	\textbf{50.40$\pm$0.12}&	\textbf{44.19$\pm$0.07} \\
       
        \bottomrule
    \end{tabular}
    }
    \caption{Multi-Source Domain adaptation on DomainNet dataset.}
    \label{tab:domainnet}
     \vspace{-10pt}
\end{table*}

\begin{table}[t]
    \centering
    \resizebox{0.55\linewidth}{!}{
    \begin{tabular}{ll|cccccc}
    \toprule
     &  Method & R$\mapsto$C	&P$\mapsto$C	&P$\mapsto$A	&A$\mapsto$C	&C$\mapsto$A	&Ave.   \\
       \midrule
    \parbox[t]{2mm}{\multirow{7}{*}{\rotatebox[origin=c]{90}{AlexNet}}}  &  S+T	    &44.6		&		44.4&	36.1&		38.8&			37.5&		40.3\\
    &DANN~\cite{ganin2016domain}	&47.2	&		44.4&	36.1&		39.8&			38.6&		41.2\\
    &ADR~\cite{saito2017adversarial}&45.0	&		38.9&	36.3&		40.0&		    37.3&		39.5\\
    &CDAN~\cite{long2018conditional}	&41.8	&		35.8&	32.0&	    34.5&			27.9&		34.4\\
    &ENT~\cite{grandvalet2005semi}	&44.9	&		41.2&	34.6&		37.8&			31.8&		38.1\\
    &MME~\cite{saito2019semi}	&\textbf{51.2}	&		47.2&	\textbf{40.7}&		43.8&			\textbf{44.7}&		45.5\\
    &Meta-MME (Ours) & 50.3	&	\textbf{48.3}&	40.3&		\textbf{44.5}&			44.5&		\textbf{45.6}\\
        \midrule
    \parbox[t]{2mm}{\multirow{4}{*}{\rotatebox[origin=c]{90}{ResNet-34}}} &    S+T &57.4&54.5&59.9&	56.2&57.6&	57.1 \\
    &ENT~\cite{grandvalet2005semi} & 62.8&61.8	&65.4&62.1&65.8&	63.6\\
    &MME~\cite{saito2019semi}	&64.9			&63.8	& 65.0	&	63.0	&		66.6&		64.7\\
    & Meta-MME (Ours)&	\textbf{65.2}&\textbf{64.5}&\textbf{66.7}&\textbf{63.3}&\textbf{67.5}&\textbf{65.4}\\
    \bottomrule
    \end{tabular}
    }
    \caption{Semi-supervised domain adaptation: Office-Home.}
    \label{tab:ssda-office-home}
\end{table}

\begin{table*}[t]
\begin{minipage}[b]{0.66\linewidth}
\centering
\resizebox{0.85\linewidth}{!}{
    \begin{tabular}{ll|ccccccc|c}
    \toprule
\cut{Backbone}&Method & R$\mapsto$C & R$\mapsto$P & P$\mapsto$C & C$\mapsto$S & S$\mapsto$P & R$\mapsto$S & P$\mapsto$R & Ave \\
\midrule
\parbox[t]{2mm}{\multirow{7}{*}{\rotatebox[origin=c]{90}{AlexNet}}}&
S+T	&	47.1	&	45.0	&	44.9&		36.4&		38.4&		33.3&		58.7&		43.4\\
&DANN\cite{ganin2016domain}	&	46.1	&	43.8&		41.0	&	36.5&		38.9&		33.4	&	57.3&		42.4\\
&ADR~\cite{saito2017adversarial}	&	46.2	&	44.4&		43.6&		36.4&		38.9&		32.4&		57.3&		42.7\\
&CDAN~\cite{long2018conditional}	&	46.8	&	45.0	&	42.3	&	29.5&		33.7	&	31.3	&	58.7&		41.0\\
&ENT~\cite{grandvalet2005semi}	&	45.5	&	42.6&		40.4&		31.1&		29.6&		29.6&		60.0&		39.8\\
&MME~\cite{saito2019semi}	&	55.6	&	49.0	&	51.7	&	39.4	&	43.0	&	37.9	&	\textbf{60.7}&		48.2\\
&Meta-MME (Ours)& \textbf{56.4}&	\textbf{50.2}	& \textbf{51.9}&	\textbf{39.6}&	\textbf{43.7}&	\textbf{38.7}&	\textbf{60.7} & \textbf{48.8}\\
\midrule
\parbox[t]{2mm}{\multirow{7}{*}{\rotatebox[origin=c]{90}{ResNet-34}}}
&S+T	&60.0	&62.2&59.4&	55.0&	59.5&50.1&73.9&	60.0\\
&DANN~\cite{ganin2016domain}&59.8&62.8&59.6&55.4&59.9&54.9&72.2&60.7\\
&ADR~\cite{saito2017adversarial}	&60.7&61.9&60.7&54.4&59.9&51.1&74.2&60.4\\
&CDAN~\cite{long2018conditional}&69.0&67.3&68.4&57.8&65.3&59.0&78.5&66.5\\
&ENT~\cite{grandvalet2005semi}	&71.0	&69.2&71.1&60.0&62.1&61.1&78.6&67.6\\
&MME~\cite{saito2019semi}	&72.2&69.7&71.7&61.8&66.8&61.9&78.5&68.9\\
&Meta-MME (Ours)& \textbf{73.5}	&\textbf{70.3}&	\textbf{72.8}&	\textbf{62.8}&	\textbf{68.0}&	\textbf{63.8}&	\textbf{79.2}&	\textbf{70.1} \\
\bottomrule
    \end{tabular}
    }
  \caption{Semi-supervised DA on DomainNet.}
    \label{tab:ssda-domainnet}
\end{minipage}\hfill
\begin{minipage}[b]{0.3\linewidth}
\includegraphics[width=1.0\linewidth]{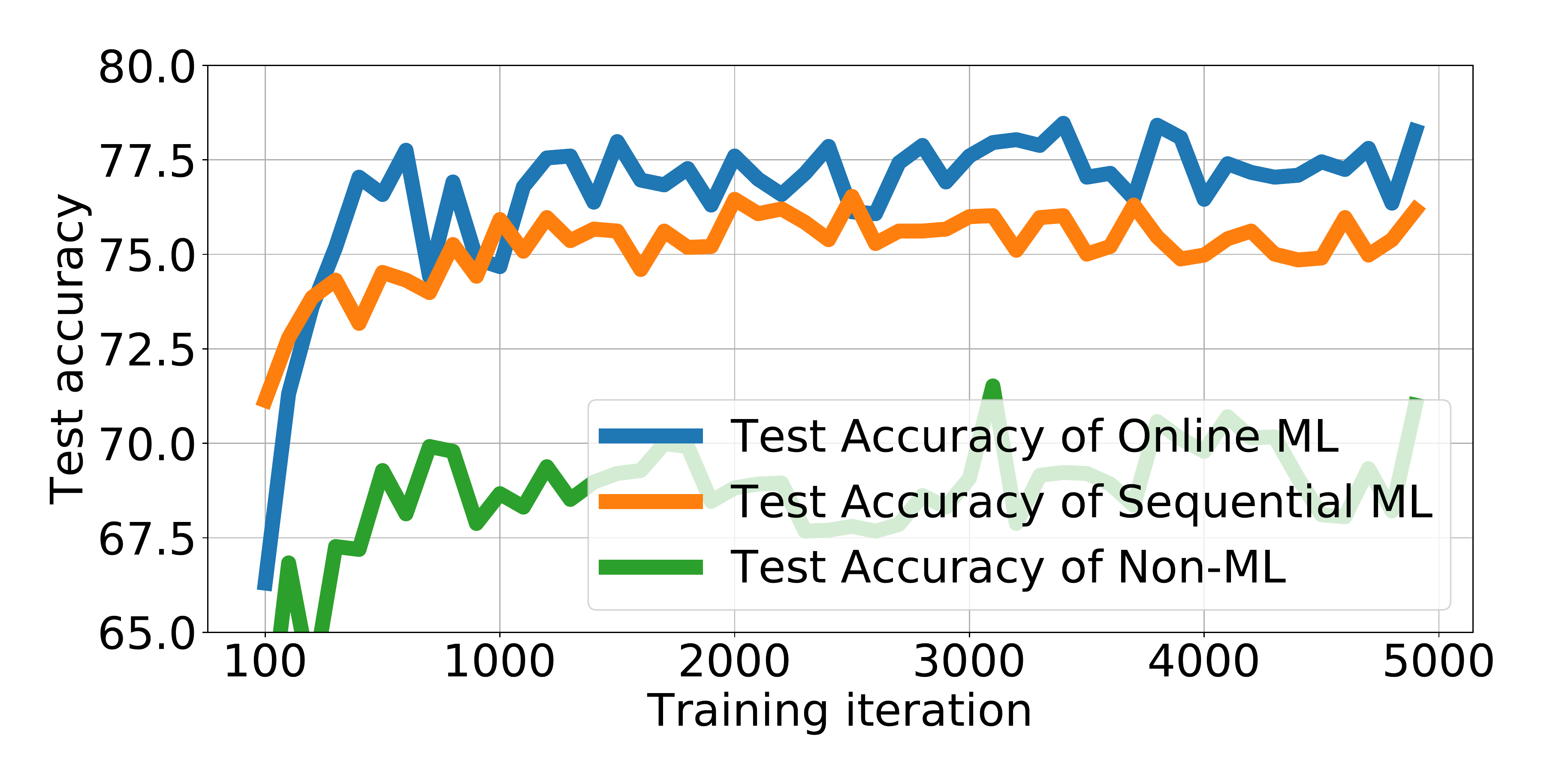}
\includegraphics[width=1.0\linewidth]{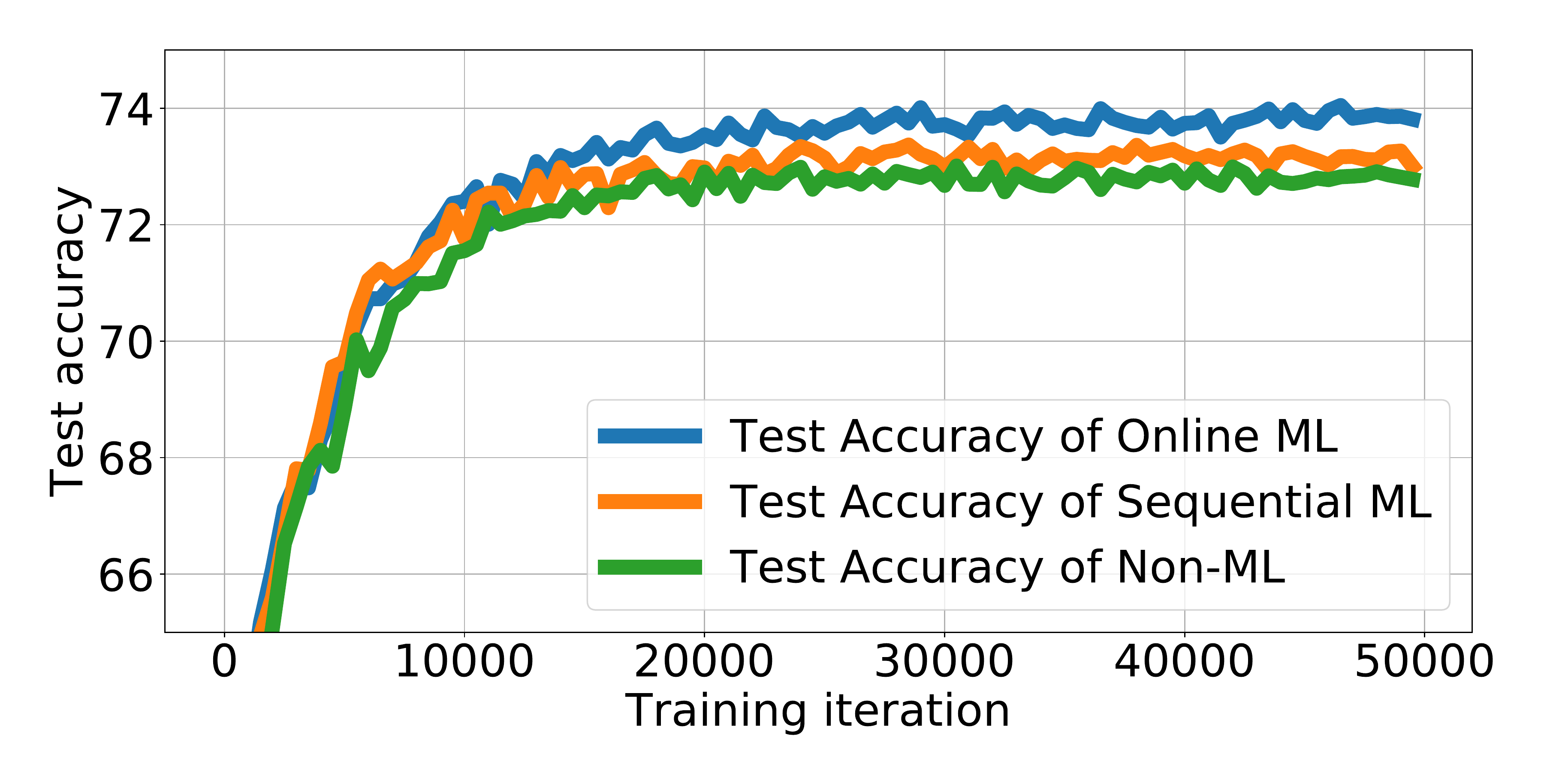}
    \captionof{figure}{Vanilla DA vs seq. and online meta (T:MSDA, B:SSDA). 
    }
    \label{fig:test-curve-iter-vs-noniter}
\end{minipage}
\vspace{-10pt}
\end{table*}

\begin{table}[tb]
   \centering
\resizebox{0.6\linewidth}{!}{
    \begin{tabular}{ll|cccc|c}
    \toprule
      Setting& Method  & C,P,S$\to$A &    A,P,S$\to$C    & A,C,S$\to$P    & A,C,P$\to$S        & Ave. \\
       \midrule
  \multirow{3}{*}{DG}  &   MetaReg~\cite{balaji2018metareg}	&83.7 &	77.2	&95.5&	70.3&	81.7 \\
    &   Epi-FCR~\cite{Li_2019_ICCV}	&82.1 &	77.0&	93.9&	73.0&	81.5\\
    &   MASF~\cite{dou2019domain}     &80.3 & 77.2&	95.0&	71.7&	81.0\\
    \midrule
  \multirow{2}{*}{DA}  &   MCD~\cite{saito2018maximum}	    &86.3 &	84.5&	97.3&	71.0&	84.8\\
    &   Meta-MCD  &87.4 &	86.2&	97.1&	78.3&	87.2\\
      \bottomrule
    \end{tabular}
    }
    \caption{Comparison between DG and DA methods on PACS.}
    \label{tab:da-vs-dg-pacs}
\end{table}

\begin{figure*}[t]
\hspace{0.2cm}
    \centering
    \begin{subfigure}[b]{0.32\linewidth}
    \centering
    \includegraphics[width=\linewidth]{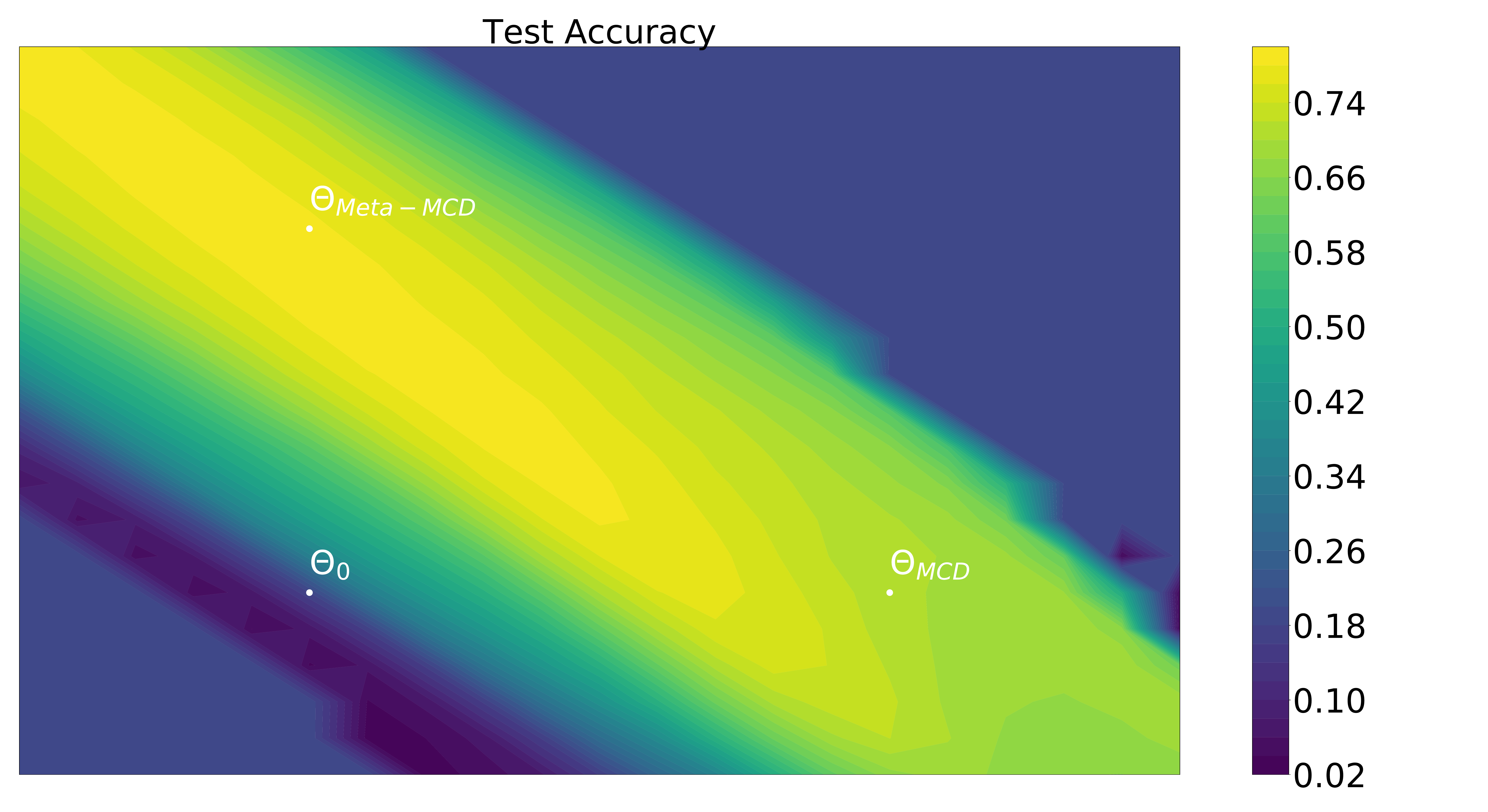}
    \caption{Test accuracy on target.}
    \label{fig:test-pacs-3points}
    \end{subfigure}
    \begin{subfigure}[b]{0.32\linewidth}
    \centering
    \includegraphics[width=\linewidth]{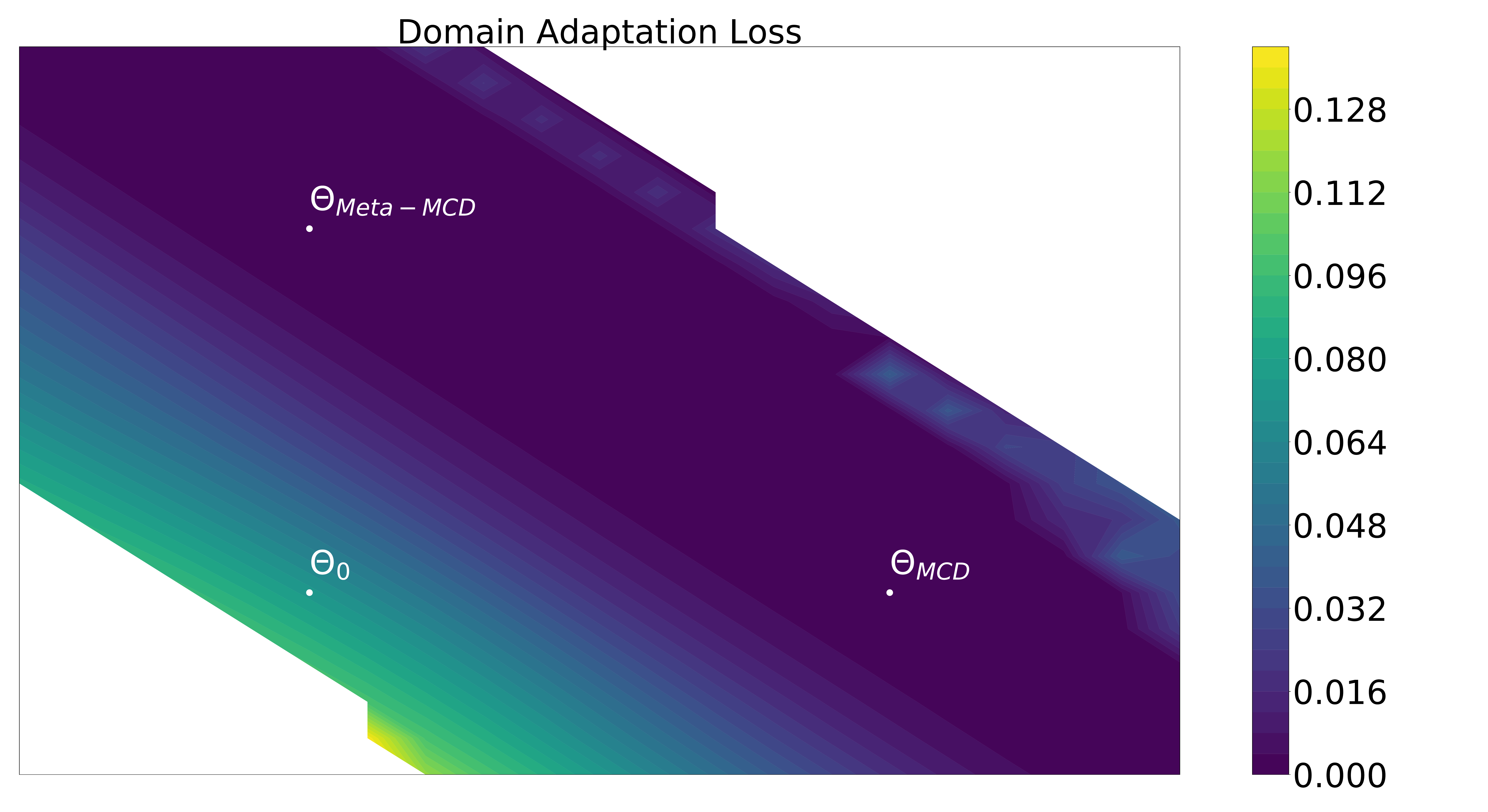}
     \caption{DA loss, $\mathcal{L}_a$.}
    \label{fig:da-loss-pacs-3points}
    \end{subfigure}
    \begin{subfigure}[b]{0.32\linewidth}
    \centering
    \includegraphics[width=\linewidth]{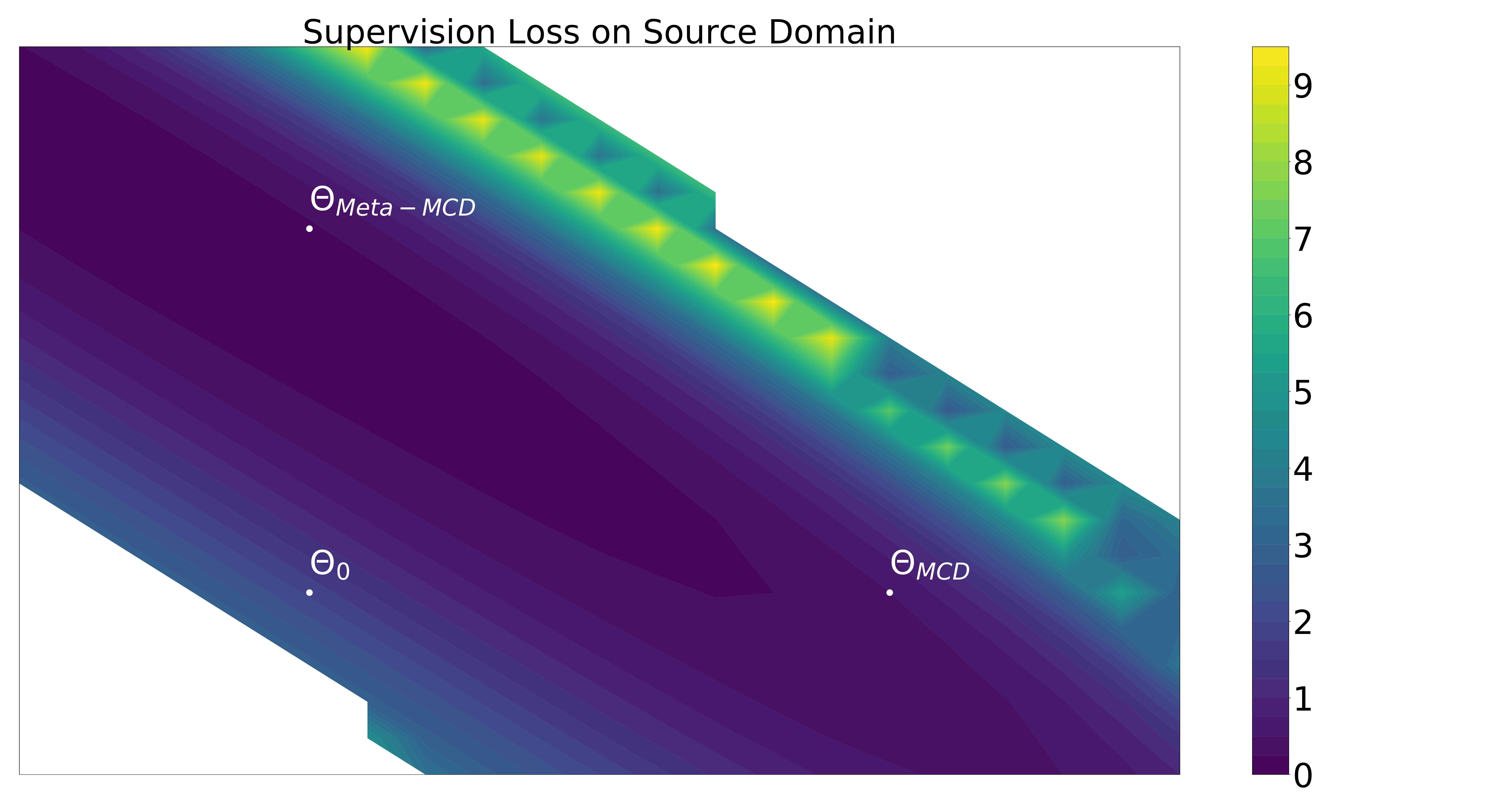}
     \caption{Supervised loss $\mathcal{L}_{\text{sup}}$.
     }
    \label{fig:train-loss-source-pacs-3points}
    \end{subfigure}
    \caption{Performance across weight space slices defined by a common initial condition $\Theta_0$ and MCD and Meta-MCD solutions ($\Theta_\text{Meta-MCD}$ and $\Theta_\text{MCD}$ respectively). MSDA PACS benchmark with Sketch target.
    \cut{$\mathcal{L}_a$ loss and $\mathcal{L}_{\text{sup}}$ loss of full data of source domain on PACS.}}
    \label{fig:contour-pacs0}
\end{figure*}

\cut{\subsubsection{Office-Caltech10 Benchmark}
\keypoint{Dataset and Settings} Office-Caltech10 benchmark contains four domains and ten intersecting categories between Office31 and Caltech256 datasets, with $\sim$2500 images in total.  We follow the experimental setting in \cite{peng2019moment} and use the ImageNet pre-trained ResNet-101 as the backbone model for fair comparison. We use M-SGD to train all  models (batch size=32, learning rate=$10^{-3}$, momentum=0.9 and weight decay=$10^{-4}$) for $1k$ iterations.
\keypoint{Results} From the results in Table~\ref{tab:office-caltech-10}, we can see that as using ResNet-101 as the backbone model, all base adaptation models achieve strong performance. MCD is best among all the baselines, and incorporating it into our framework, Meta-MCD provides a slight increase in performance. Overall the performance on this benchmark is rather saturated, so it is difficult for Meta-MCD to provide significant improvement. Nevertheless, Meta-MCD sets a state of the art result here.}

\vspace{0.2cm}\keypoint{Office-Home:}
\keypoint{Dataset and Settings} Office-Home was initially proposed for the single-source domain adaptation, containing $\approx15,500$ images from four domains `artistic', `clip art', `product' and `real-world' with 65 different categories. We follow the setting in~\cite{chang2019domain} and use  ImageNet pretrained ResNet-50 as our backbone. We train all models with M-SGD (batch size=32, learning rate=$10^{-3}$, momentum=0.9 and weight decay=$10^{-4}$) for $3k$ iterations. 
\keypoint{Results} From  Table~\ref{tab:office-home}, we see that MCD achieves the best performance among the baselines. Applying our meta-learning framework improves both baselines by a small amount, and Meta-MCD achieves state-of-the-art performance on this benchmark.

\vspace{0.2cm}\keypoint{DomainNet:}
\keypoint{Dataset} DomainNet is a recently benchmark \cite{peng2019moment} for multi-source domain adaptation in object recognition. It is the largest-scale DA benchmark so far, with $\approx0.6$m images across six domains and 345 categories.
\keypoint{Settings} We follow the official train/test split protocol \cite{peng2019moment}\footnote{Other settings such as optimizer, iterations and data augmentation are not clearly stated in~\cite{peng2019moment}, making it hard to replicate their results.}. Various feature extraction backbones were used in the original paper \cite{peng2019moment}, making it hard to compare results. We use ImageNet pre-trained ResNet-18 and ResNet-34 for our own implementations to facilitate direct comparison. We use M-SGD to train all the competitors (batch size=32, learning rate=0.001, momentum=0.9, weight decay=0.0001) for $10k$ iterations\footnote{We tried training with up to $50k$, and found it did not lead to clear improvement. So, we train all models for $10k$ iterations to minimise cost.}. We re-train the model three times to generate standard deviations.
\keypoint{Results} From the results in Table~\ref{tab:domainnet}, we can see that: (i) The top group of results from~\cite{peng2019moment} show that the dataset is a much more challenging domain adaptation benchmark than previous ones. Most existing domain adaptation methods (typically tuned on small-scale benchmarks) fail to improve over the source-only baseline according to the results in~\cite{peng2019moment}. (ii) The middle group of ResNet-18 results show that our MCD experiment achieves comparable results to those in~\cite{peng2019moment}. (iii) Our Meta-MCD and Meta-DANN methods provide a small but consistent improvement over the corresponding MCD and DANN baselines for both ResNet-18 and ResNet-34 backbones. While the improvement margins are relatively small, this is a significant outcome as the results show that the base DA methods already struggle to make a large improvement over the source-only baseline when using ResNet-18/34; and also the multi-run standard deviation is small compared to the margins. (iv) Overall our Meta-MCD achieves state-of-the-art performance on the benchmark by a small margin. 

\subsection{Semi-Supervised Domain Adaptation}
\subsubsection{Office-Home:}
\keypoint{Setting} We follow the setting in~\cite{saito2019semi}. We focus on 3-shot learning in the target domain (three annotated examples only per category), and focus on the five most difficult source-target domain pairs. We use the ImageNet pretrained AlexNet and ResNet-34 as  backbone models. We train all the models with M-SGD, with batch size 24 for labelled source and target domains and 48 for the unlabelled target as in~\cite{saito2019semi}, learning rate is $10^{-2}$ and $10^{-3}$ for the fully-connected and the rest trainable layers. We also use horizontal-flipping and random-cropping data augmentation for training images.
\keypoint{Results} From the results in Table~\ref{tab:ssda-office-home}, we can see that our Meta-MME does not impact  performance on AlexNet. However, for a modern ResNet-34 architecture,  Meta-MME provides a visible $\sim0.8\%$ accuracy gain over the MME baseline, which results in the state-of-the-art performance of SSDA on this benchmark.

\vspace{0.2cm}
\keypoint{DomainNet:}
\keypoint{Settings} We evaluate DomainNet for 1-1 few-shot domain adaptation as in~\cite{saito2019semi}. We evaluate  both AlexNet and modern ResNet-34 backbones, and apply our meta-learning method on MME. As per~\cite{saito2019semi}, we train our models using M-SGD where the initial learning rate is 0.01 for the fully-connected layers and 0.001 for the rest of trainable layers. During the training we use the annealing strategy in~\cite{ganin2016domain} to decay the learning rate, and use the same batch size as~\cite{saito2019semi}.
\keypoint{Results} From the results in Table~\ref{tab:ssda-domainnet}, we can see our Meta-MME improves on the accuracy of the base MME algorithm in all pairwise transfer choices, and also for both backbones. These results show the consistent effectiveness of our method, as well as improving state-of-the-art for DomainNet SSDA.

\subsection{Further Analysis}\label{sec:further}

\keypoint{Discussion} \doublecheck{Our final online algorithm can be understood as performing DA with periodic meta-updates that adjust parameters to optimize the impact of the following DA steps. From the perspective of any given DA step, the role of the preceding meta-update is to tune its initial condition.}\\
\keypoint{Non-Meta vs Sequential vs Online Meta} 
This work is the first to propose meta-learning to improve domain adaptation, and in particular to contribute an efficient and effective online meta-learning algorithm for initial condition training. Exact meta learning is intractable to compare. However, this section we compare our online meta update with the alternative sequential approximation, and non-meta alternatives for both MSDA and SSDA using A,C,P$\to$S and R$\to$C as examples. 
For fair comparison, we control the number of meta-updates ($\operatorname{UpdateIC}$) and base DA updates available to both sequential and online meta-learning methods to the same amount. Figure~\ref{fig:test-curve-iter-vs-noniter} shows that: (1) Sequential meta-learning method already improves the performance on the target domain comparing to vanilla domain adaptation, which confirms the potential for improvement by refining model initialization. (2) The sequential strategy has a slight advantage early in DA training, which makes sense, as all meta-updates occur in advance. But overall our online method that interleaves meta-updates and DA updates leads to higher test accuracy.\\
\keypoint{Computational Cost} Our Meta-DA imposes only a small computational overhead over the base DA algorithm. For example, comparing Meta-MCD and MCD on ResNet-34 DomainNet, the time per iteration is $2.96$s vs $2.49$s respectively\footnote{Using GeForce RTX 2080 GPU. Xeon Gold 6130 CPU @ 2.10GHz.}. \\
\keypoint{Weight-Space Illustration} To investigate our method's mechanism, we train MCD and Meta-MCD from a common initial condition on MSDA PACS when `Sketch' is the target domain. We use the initial condition $\Theta_0$ and two different solutions ($\Theta_\text{Meta-MCD}$ and $\Theta_\text{MCD}$) to define a plane in weight-space and colour it according to the performance at each point. We can see from Figure~\ref{fig:contour-pacs0}(a) that Meta-MCD finds a solution with greater test accuracy. Figures~\ref{fig:contour-pacs0}(b) and (c) break down the training loss components. We can see that, in this slice, both methods managed to minimize MCD's adaptation (classifier discrepancy) loss $\mathcal{L}_a$ adequately, but MCD failed to minimize the supervised loss as well as Meta-MCD ($\Theta_\text{Meta-MCD}$ is closer to the minima than $\Theta_\text{MCD}$). Note that both methods were trained to convergence in generating these solutions. This suggests that Meta-MCD's meta-optimization step using meta-train/meta-test splits materially benefits the optimization dynamics of the downstream MSDA task.\\
\keypoint{Model Agnostic} We emphasize that, although we focused on DANN, MCD and MME, our MetaDA framework can apply to any base DA algorithm. Supplementary~\ref{sec:additional}  shows some results for  JiGen and M$^3$SDA algorithms.\\
\keypoint{Comparison between DA and DG methods}
As a highly related topical problem to domain adaptation, domain generalization assumes no access to the target domain data during the training. DA and DG methods are rarely directly compared. Now we compare our Meta-MCD and MCD with some state of the art DG methods on PACS as shown in Table~\ref{tab:da-vs-dg-pacs}. From the results, we can see that generally DA methods outperform the DG methods with a noticeable margin, which is expected as DA methods `see' the target domain data at training.

\section{Conclusion\label{sec:conc}}
We proposed a meta-learning pipeline to improve domain adaptation by initial condition optimization. Our online shortest-path solution is efficient and effective, and provides  a consistent boost to several domain adaptation algorithms, improving state of the art in both multi-source and semi-supervised settings. Our approach is agnostic to the base adaptation method, and can potentially be used to improve many DA algorithms that fit a very general template. In future we aim to meta-learn other DA hyper-parameters beyond initial conditions.

\newpage
\clearpage
{\small
\bibliographystyle{splncs04}
\bibliography{egbib}
}
%

\clearpage
\appendix

\section{Short-Path Gradient Descent}

Optimizing Eq.~\ref{eq:meta-learning} naively by Algorithm~\ref{alg:bilevel-ml} would be costly and ineffective. It is costly because in the case of domain adaptation (unlike for example, few-shot learning \cite{finn2017model}, the inner loop requires many iterations). So back-propagating through the whole optimization path to update the initial $\Theta$ in the outer loop will produce multiple high-order gradients. For example, if the inner loop applies $j$ iterations, we will have


\begin{equation}\label{eq:inner-two-steps}
\begin{aligned}
 \Theta_{}^{(1)} & = \Theta_{}^{} - \alpha \nabla_{\Theta_{}^{(0)}} \mathcal{L}_{\text{uda}}(.) \\
     \dots\\
    \Theta_{}^{(j)} & = \Theta_{}^{(j-1)} - \alpha\nabla_{\Theta_{}^{(j-1)}}\mathcal{L}_{\text{uda}}(.)
\end{aligned}
\end{equation}
then the outer loop will update the initial condition as
 \begin{equation}\label{eq:outer-orig}
\begin{aligned}
      \Theta^* = \Theta - \alpha\overbrace{\nabla_{\Theta}\mathcal{L}_{\text{sup}}(\Theta_{}^{(j)} , \mathcal{D}_{\text{val}})}^{\text{Meta Gradient}}
\end{aligned}
\end{equation}
where higher-order gradient will be required for all items  $\scriptstyle \nabla_{\Theta_{}^{(0)}}\mathcal{L}_{\text{uda}}(.), \dots,\nabla_{\Theta_{}^{(j-1)}}\mathcal{L}_{\text{uda}}(.)$ in the update of Eq.~\ref{eq:outer-orig}.

One intuitive way of eliminating higher-order gradients for computing Eq.~\ref{eq:outer-orig} is making $\nabla_{\Theta_{}^{(0)}}\mathcal{L}_{\text{uda}}(.),\dots ,\nabla_{\Theta_{}^{(j-1)}}\mathcal{L}_{\text{uda}}(.)$ constant during the optimization. Then, Eq.~\ref{eq:outer-orig} is equivalent to
\begin{equation}\label{eq:outer-const}
\begin{aligned}
      \Theta^* = \Theta - \alpha \overbrace{\nabla_{\Theta^{(j)}}\mathcal{L}_{\text{sup}}(\Theta^{(j)} , \mathcal{D}_\text{val})}^{\text{First-order Meta Gradient}}
\end{aligned}
\end{equation}
However, in order to compute Eq.~\ref{eq:outer-const}, one still needs to store the optimization path of Eq.~\ref{eq:inner-two-steps} in memory and back-propagate through it to optimize $\Theta$, which requires high computational load.
Therefore, we propose a practical solution an iterative meta-learning algorithm to iteratively optimize the model parameters during training.

\keypoint{Shortest Path Optimization}
To obtain the meta gradient in Eq.~\ref{eq:outer-const} in a more efficient way, we propose a more scalable and efficient meta-learning method using shortest-path gradient (S-P.G.)~\cite{nichol2018first}. 
Before the optimization of Eq.~\ref{eq:inner-two-steps}, we copy the parameters $\Theta$ as $\tilde{\Theta}^{(0)}$ and use $\tilde{\Theta}^{(0)}$ in the inner-level algorithm. 
\begin{equation}\label{eq:inner-copy}
    \begin{aligned}
    \tilde{\Theta}^{(j)}=
    \begin{cases} 
    \tilde{\Theta}^{(0)} - \alpha \nabla_{\Theta^{(0)}}\mathcal{L}_{\text{uda}}(\tilde{\Theta}^{(0)}, \mathcal{D}_\text{tr}), \\
    \dots \\
    \tilde{\Theta}^{(j-1)} -\alpha \nabla_{\Theta^{(0)}}\mathcal{L}_{\text{uda}}(\tilde{\Theta}^{(j-1)}, \mathcal{D}_\text{tr})
    \end{cases}
    \end{aligned}
\end{equation}
then, after finishing the optimization in Eq.~\ref{eq:inner-copy}, we can get the shortest-path gradient between two items $\tilde{\Theta_i}^{(j)}$ and $\Theta_i$.

\begin{equation}\label{eq:short-gradient}
    \begin{aligned}
    \nabla_{\Theta}^{\text{short}} = \Theta-  \tilde{\Theta}^{(j)}
    \end{aligned}
\end{equation}

Different from Eq.~\ref{eq:outer-const}, we use this shortest-path gradient $\nabla_{\Theta}^{\text{short}}$ and initial parameter $\Theta$ to compute  $\mathcal{L}_{\text{sup}}(.)$ as,
\begin{equation}\label{eq:loss-sup-short}
    \begin{aligned}
     \mathcal{L}_{\text{sup}}(\Theta_i - \nabla_{\Theta_i}^{\text{short}} , \mathcal{D}_{\text{val}})
    \end{aligned}
\end{equation}
Then, one-step meta update of Eq.~\ref{eq:loss-sup-short} will be,
\begin{equation}\label{eq:outer-short}
\begin{aligned}
      \Theta_i^* &= \Theta_i - \alpha \nabla_{\Theta_i}\mathcal{L}_{\text{sup}}(\Theta_i - \nabla_{\Theta_i}^{\text{short}} , \mathcal{D}_{\text{val}}) \\
      &=\Theta_i - \alpha \nabla_{\Theta_i-\nabla_{\Theta_i}^{\text{short}}}\mathcal{L}_{\text{sup}}(\Theta_i - \nabla_{\Theta_i}^{\text{short}} , \mathcal{D}_{\text{val}}) \\
      &= \Theta_i - \alpha \nabla_{\tilde{\Theta_i}^{(j)}}\mathcal{L}_{\text{sup}}(\tilde{\Theta_i}^{(j)} , \mathcal{D}_{\text{val}})
\end{aligned}
\end{equation}

\noindent\textbf{Effectiveness:} We can see that one update of Eq.~\ref{eq:outer-short} corresponds to that of Eq.~\ref{eq:outer-const}, which proves that using shortest-path optimization has the equivalent effectiveness to the first-order meta optimization. \textbf{Scalability/Efficiency:} The computation memory of the first-order meta-learning increases linearly with the inner-loop update steps, which is constrained by the total GPU memory. However, for the shortest-path optimization, storing the optimization graph is no longer necessary, which makes it scalable and efficient. We also experimentally evaluate that one step shortest-path optimization is 7x faster than one-step first-order meta optimization in our setting.
The overall algorithm flow is shown in Algorithm~\ref{alg:bilevel-ml-short}.

\section{Additional Illustrative Schematics}
\begin{figure}[t]
    \centering
    \begin{subfigure}{0.49\linewidth}
    \includegraphics[width=1.0\linewidth]{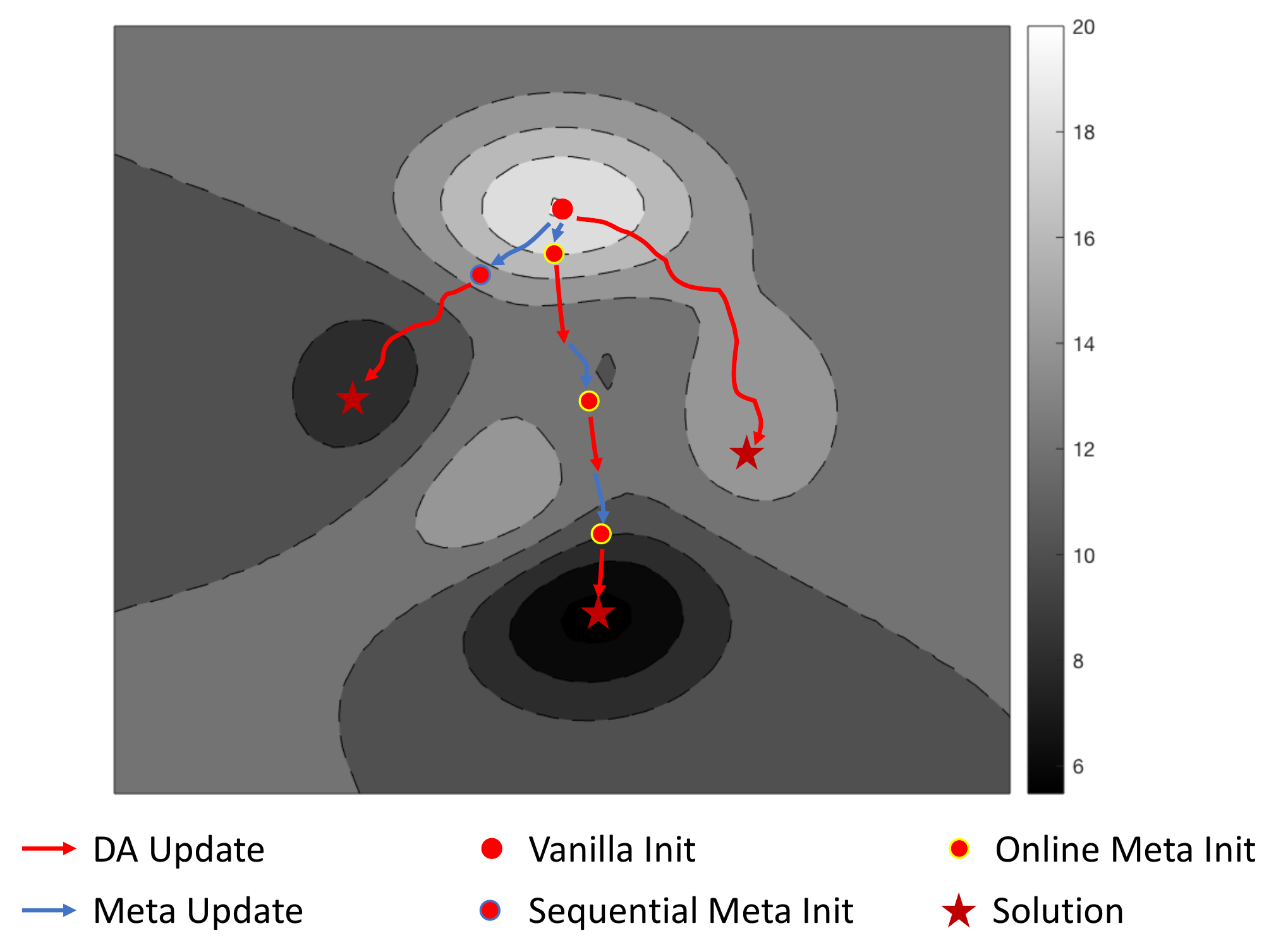}
    \end{subfigure}
    \begin{subfigure}{0.49\linewidth}
    \includegraphics[width=1.0\linewidth]{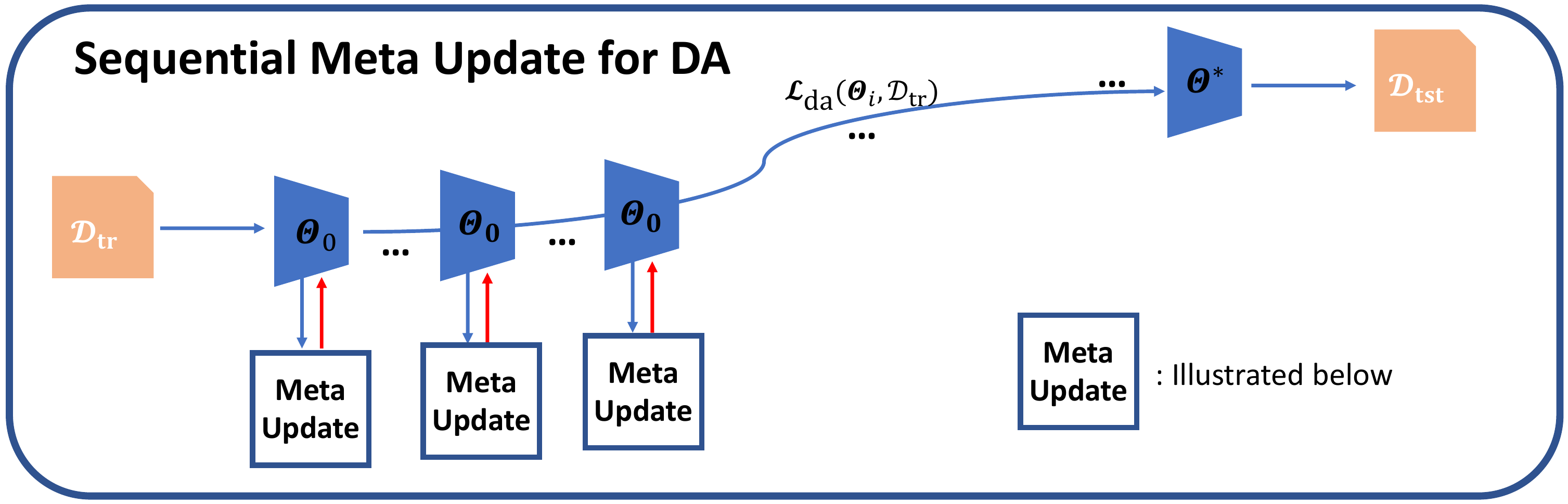}
    \includegraphics[width=1.\linewidth]{figure/4-3conceptsv2.pdf}
    \end{subfigure}
    \caption{Illustrative schematics of sequential and online meta domain adaptation. Left: Optimization paths of different approaches on domain adaptation loss (shading). 
    (Solid line) Vanilla gradient descent on a DA objective from a fixed start point. 
    (Multi-segment line) Online meta-learning iterates meta and gradient descent updates.
    (Two segment line) Sequential meta-learning provides an alternative approximation: update initial condition, then perform gradient descent. 
    Right: (Top) Sequential meta-learning performs meta updates and DA updates sequentially. (Bottom) Online meta-learning alternates between meta-optimization and domain adaptation.
    }
        \label{fig:illustration-appendix}  
\end{figure}

To better explain the contrast between our online meta-learning domain adaptation approach with the sequential meta-learning approach, we add a schematic illustration in Figure~\ref{fig:illustration-appendix}. The main difference between sequential and online meta-learning approaches is how do we distribute the meta and DA updates. Sequential meta-learning approach performs meta updates and DA updates sequentially. And online meta-learning conducts the alternative meta and DA updates throughout the whole training procedure.

\section{Additional Experiments}\label{sec:additional}

\begin{figure}
    \centering
    \includegraphics[width=0.45\linewidth]{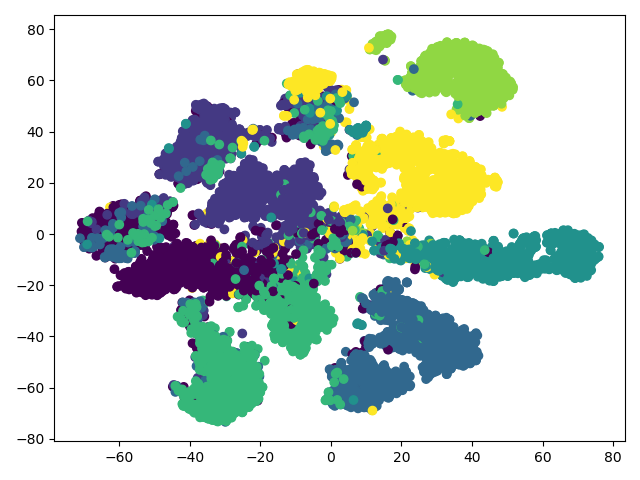}
    \includegraphics[width=0.45\linewidth]{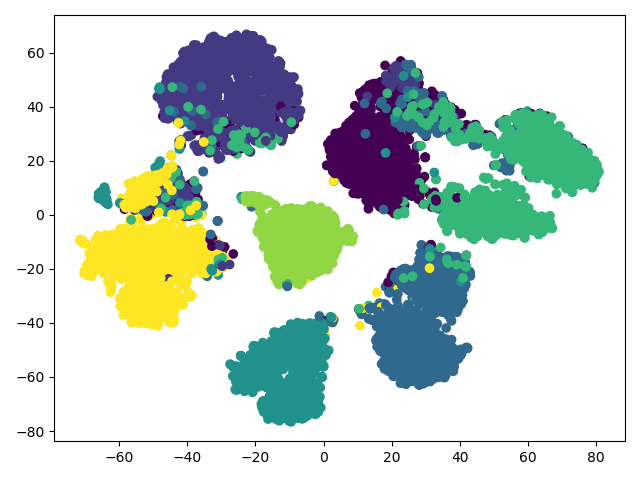}
    \caption{t-SNE~\cite{maaten2008visualizing} visualization of learned MCD (left) and Meta-MCD (right) features on PACS (sketch as target domain). Different colors indicate different categories.}
    \label{fig:vis-feature}
\end{figure}

\keypoint{Visualization of the Learned Features}
We visualize the learned features of MCD and Meta-MCD on PACS when sketch is the target domain as shown in Figure~\ref{fig:vis-feature}. We can see that both MCD and Meta-MCD can learn discriminative features. However, the features learned by Meta-MCD is more separable than vanilla MCD. This explains why our Meta-MCD performs better than the vanilla MCD method.

\begin{table}[t]
    \centering
    \resizebox{.7\linewidth}{!}{
    \begin{tabular}{c|ccc}
    \toprule
    Method     &  Meta-MCD ($S$=3) &Meta-MCD ($S$=5) & Meta-MCD ($S$=10) \\
     DomainNet (ave.)    & 41.02 & 40.98 & 40.93 \\
     \bottomrule
    \end{tabular}
    }
    \caption{MetaDA is insensitive to the update ratio hyperparameter $S$ --  Results for MSDA ResNet-18 performance on DomainNet. }
    \label{tab:vary-s}
\end{table}

\keypoint{Effect of varying $S$}
Our online meta-learning method has iteration hyper-parameters $S$ and $J$. We fix $J=1$ throughout, and analyze the effect of varying $S$ here using the DomainNet MSDA experiment with ResNet-18. The result in Table~\ref{tab:vary-s} shows that MetaDA is rather insensitive to this hyperparameter.

\keypoint{Varying the Number of Source Domains in MSDA} For multi-source DA, the performance of both Meta-DA and the baselines is expected to drop with fewer sources (same for SSDA if fewer labeled target domain points). To disentangle the impact of the number of sources for baseline vs Meta-DA we compare MSDA by Meta-MCD on PACS with 2 vs 3 sources. The results for Meta-MCD vs vanilla MCD are 82.30\% vs 80.07\% (two source, gap 2.23\%) and 87.24\% vs 84.79\% (three source, gap 2.45\%). Meta-DA margin is similar with reduction of domains. Most difference is accounted for by the impact on the base DA algorithm.

\keypoint{Other base DA methods}
Besides the base DA methods evaluated in the main paper (DANN, MCD and MME), our method is applicable to any base domain adaptation method. We use the  published code of JiGen\footnote{https://github.com/fmcarlucci/JigenDG} and M$^3$SDA\footnote{https://github.com/VisionLearningGroup/VisionLearningGroup.github.io}, and further apply our Meta-DA on the existing code. The results are shown in Table~\ref{tab:meta-jigen} and~\ref{tab:meta-msda}. From the results, we can see that our Meta-JiGen and Meta-M$^3$SDA-$\beta$ improves over the base methods by 3.42\% and 1.2\% accuracy respectively, which confirms our Meta-DA's generality. The reason we excluded these from the main results is that: (i) Re-running JiGen's published code on our compute environment failed to replicate their published numbers. (ii) M$^3$SDA as a base algorithm is very slow to run comprehensive experiments on. Nevertheless, these results provide further evidence that Meta-DA can be a useful module going forward to plug in and improve future new base DA methods as well as those evaluated here.

\keypoint{Initialization Dependence of Domain Adaptation} One may not think of domain adaptation as being sensitive to initial condition, but given the lack of target domain supervision to guide learning, different initialization can lead to a significant difference in accuracy. To illustrate this we re-ran MCD-based DA on PACS with sketch target using different initializations. From the results in Tab~\ref{tab:diff-init}, we can see that both different classic initialization heuristics, and simple perturbation of a given initial condition with noise can lead to significant differences in final performance. This confirms that studying methods for tuning initialization provide a valid research direction for advancing DA performance.

\begin{table}[tb]
   \centering
\resizebox{0.7\linewidth}{!}{
    \begin{tabular}{l|cccc|c}
    \toprule
       Method  & C,P,S$\to$A &    A,P,S$\to$C    & A,C,S$\to$P    & A,C,P$\to$S    & Ave. \\
       \midrule
        JiGen~\cite{carlucci2019domain} & 84.88 & 81.07 & 97.96 & 79.05 & 85.74 \\
      JiGen*   & 81.54	&85.88&	97.25	&68.21	&83.22\\
      Meta-JiGen & 85.21	& 86.13 &	97.31 &	77.91&	~~86.64 (+3.42)\\
      \bottomrule
    \end{tabular}
    }
    \caption{Test accuracy on PACS. * our run.}
    \label{tab:meta-jigen}
\end{table}

\begin{table}[tb]
\centering
    \resizebox{0.8\linewidth}{!}{
    \begin{tabular}{c|ccccc|c}
    \toprule
       \multirow{2}{*}{Method}  & mt,up,sv,sy &    mm,up,sv,sy    &mt,mm,sv,sy    &mt,mm,up,sy    &mt,mm,up,sv& \multirow{2}{*}{Ave.} \\
       & $\to$mm &    $\to$mt    &$\to$up    &$\to$sv    &$\to$sy&  \\
       \midrule
      M$^3$SDA-$\beta$~\cite{peng2019moment}   & 72.82 &    98.43    &96.14    &81.32    &89.58    &87.65\\
      Meta-M$^3$SDA-$\beta$ & 71.73    &98.79    &97.80    &84.81    &91.12    &~~88.85 (+1.2) \\
      \bottomrule
    \end{tabular}
    }
    \caption{Test accuracy on Digit-Five.}
    \label{tab:meta-msda}
\end{table}

\begin{table}[tb]
    \centering
    \resizebox{0.7\linewidth}{!}{
    \begin{tabular}{c|cccc}
    \toprule
    \multirow{2}{*}{Classifier Init}    & Kaiming $\mathcal{U}$ & Xavier $\mathcal{U}$    &      Kaiming $\mathcal{N}$  & Xavier $\mathcal{N}$  \\
    & 74.49 & 73.02 & 64.27 & 73.66\\
       \midrule
   \multirow{2}{*}{Feat. Extr. Init}    & No perturb & + $\epsilon\in\mathcal{N}(0, 0.01)$  &   + $\epsilon\in\mathcal{N}(0, 0.02)$     & + $\epsilon\in\mathcal{N}(0, 0.03)$   \\
    & 74.49 & 71.85 & 59.99 & 52.18 \\
      \bottomrule
    \end{tabular}
    }
    \caption{Test accuracy of MCD on PACS (sketch) with different initialization.}
    \label{tab:diff-init}
\end{table}

\end{document}